\documentclass[journal]{IEEEtran}
\usepackage{amsmath}
\usepackage{graphicx}
\usepackage{algorithm}
\usepackage{algorithmic}
\usepackage{multirow}
\usepackage{color,xcolor}
\newcommand{\tabincell}[2]{\begin{tabular}{@{}#1@{}}#2\end{tabular}}  
\ifCLASSINFOpdf
\else
\fi
\begin{document}
\title{Disentangling Semantic-to-visual Confusion \\ for Zero-shot Learning}

\author{Zihan~Ye\thanks{Z. Ye is with Suzhou University of Science and Technology (E-mail: zihhye@outlook.com).},~\IEEEmembership{Student Member,~IEEE}, Fuyuan~Hu$^\dagger$,~\IEEEmembership{Member,~IEEE}, Fan~Lyu\thanks{F. Lyu is with Tianjin University.}, Linyan Li\thanks{L. Li is with Suzhou Institute of Trade \& Commerce.}, and~Kaizhu~Huang$^\dagger$\thanks{$^\dagger$ Corresponding authors: Prof. Hu is with Suzhou University of Science and Technology, Suzhou, 215009 China (E-mail: fuyuanhu@mail.usts.edu.cn). Prof. Huang is with the Department of Electrical and Electronic Engineering in Xi’an Jiaotong-Liverpool University, Suzhou, 215123 China (E-mail: Kaizhu.Huang@xjtlu.edu.cn).},~\IEEEmembership{Senior Member,~IEEE}}

\markboth{IEEE Transactions on Multimedia}%
{Shell \MakeLowercase{\textit{et al.}}: Bare Demo of IEEEtran.cls for IEEE Journals}

\maketitle

\begin{abstract}
Using generative models to synthesize visual features from semantic distribution is one of the most popular solutions to ZSL image classification in recent years.
The triplet loss (TL) is popularly used to generate realistic visual distributions from semantics by automatically searching discriminative representations.
However, the traditional TL cannot search reliable unseen disentangled representations  due to the unavailability of unseen classes in ZSL.
 To alleviate this drawback,  we propose in this work a multi-modal triplet loss (MMTL) which utilizes multi-modal information to search a \textit{disentangled} representation space.
As such, all classes can interplay which can benefit learning disentangled class representations in the searched space.
Furthermore, we develop a novel model called Disentangling Class Representation Generative Adversarial Network (DCR-GAN) focusing on exploiting the disentangled representations in training, feature synthesis, and final recognition stages.
Benefiting from the disentangled representations, DCR-GAN could fit a more realistic distribution over both seen and unseen features.
Extensive experiments show that our proposed model can lead to superior performance to the state-of-the-arts on four benchmark datasets. Our code is available at  \textcolor{red}{\emph{https://github.com/FouriYe/DCRGAN-TMM}}.

\end{abstract}


\IEEEpeerreviewmaketitle
\begin{figure}[htbp]
	\centering
	\includegraphics[width=0.95\linewidth]{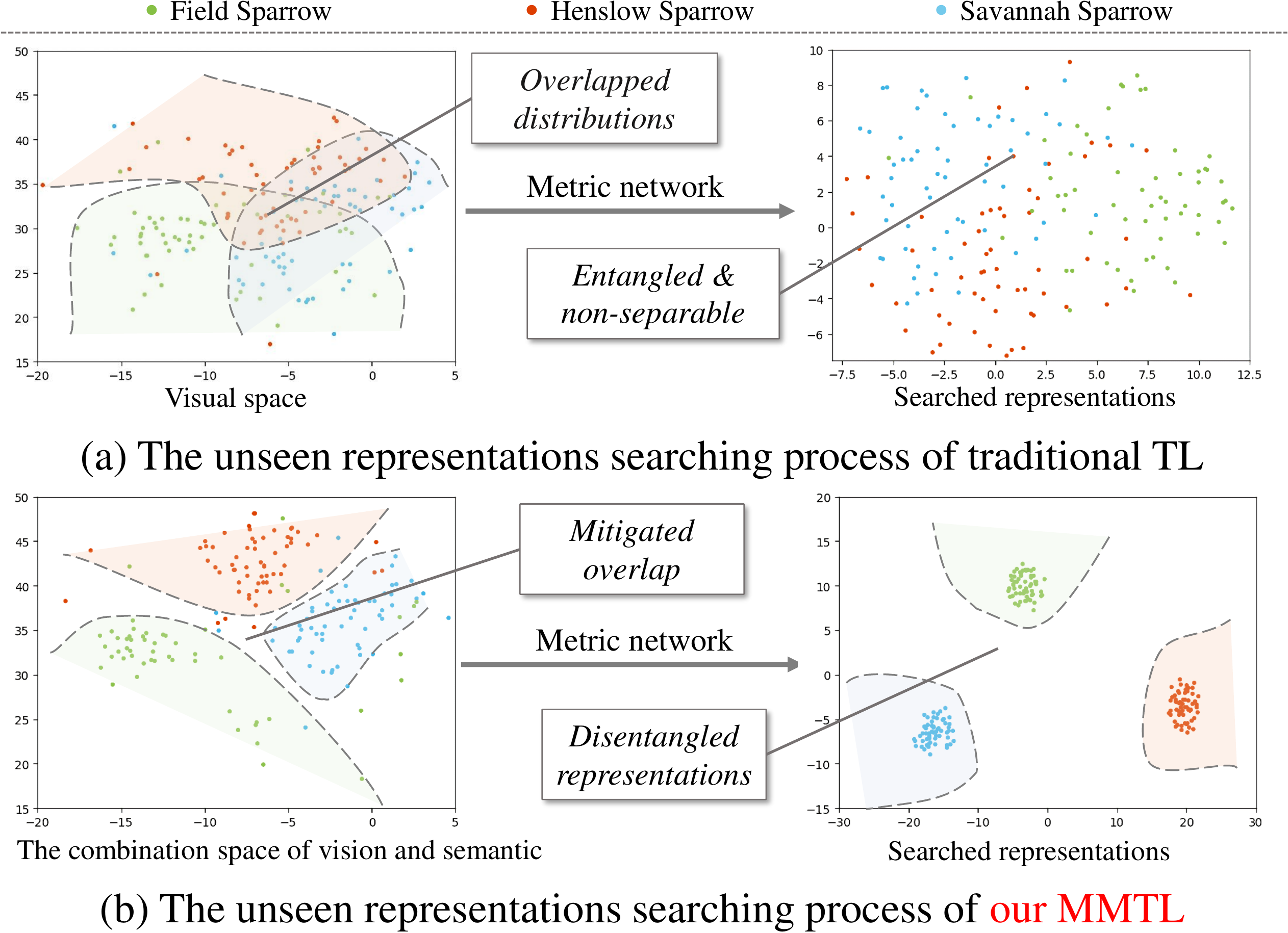}
	\caption{Comparison between the traditional triplet loss (TL) and our multi-modal triplet loss (MMTL) on three unseen bird classes of CUB.
	Due to overlapped unseen visual distributions, searched representations by the tradition TL are too entangled.
	However, MMTL mitigates the overlapped unseen visual feature problem by combining visual and semantic features, and searches disentangled unseen representations consequently.}
	\label{fig_banner}
\end{figure}
\begin{figure*}[htbp]
	\centering
	\includegraphics[width=0.90\linewidth]{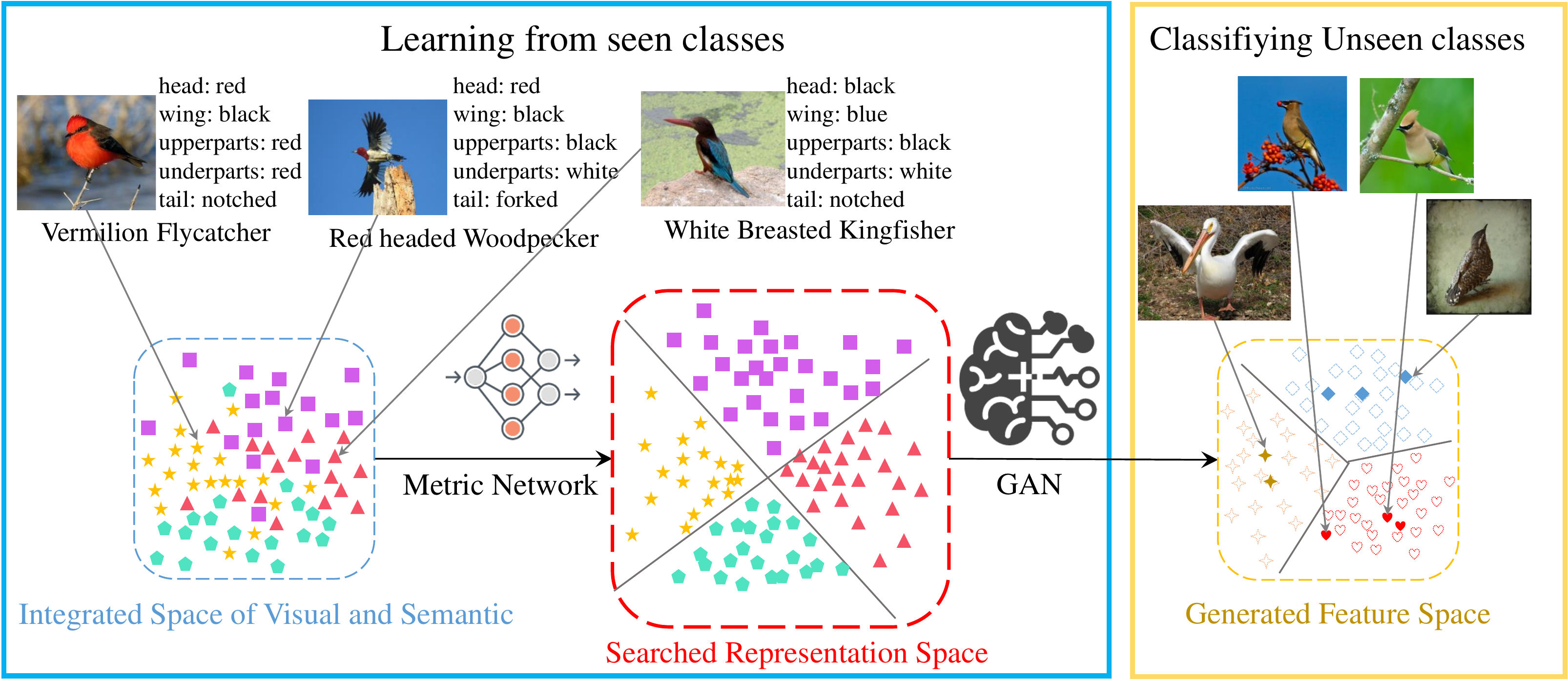}
	\caption{Illustration of our proposed DCR-GAN approach.
		We develop the metric network to construct a clear class representation space SRS by  the combination of visual and semantic spaces.
		Unseen class representation $R(\mathbf{a})$ is used to synthesize unseen visual features $\mathbf{\tilde{x}}$.
		As a result, we can recognize unseen class in the generated feature space.}
	\label{fig:approach}
\end{figure*}

\section{Introduction}

Classical pattern recognition classifies images into categories only seen in the training stage~\cite{he2016deep,lyu2019attend,zhang2017pick}.
In contrast, zero-shot learning (ZSL), one of the most active research topics in multimedia, aims at exploring unseen categories, which has recently drawn much attention~\cite{lampert2014attribute,xian2017zero,akata2016label,lampert2009learning,reed2016learning,zhu2018a,li2018discriminative,song2018selective,jiang2018learning}.
Furthermore, Chao et al. propose the generalized zero-shot learning (GZSL)~\cite{chao2016an} in a more practical scenario.
Different from ZSL, GZSL intends to recognize both seen and unseen classes during test time.
Since ZSL/GZSL does not require a vast amount of  new data, ZSL models could be utilized as an imitative solution in crucial and life-saving situations, e.g. current COVID-19 literature search~\cite{sean2020SLEDGE}, autonomous driving planning~\cite{filos2020can,yu2020zero}.

To conduct zero-shot classification, researchers usually engage intermediate semantic features to bridge the gap from seen to unseen classes.
Intermediate semantic features have many alternatives, including attribute annotations~\cite{lampert2009learning}, text representations from online text corpora~\cite{zhu2018a}, and even gaze embedding~\cite{karessli2017gaze}.
Based on these semantic features, researchers have explored two dominated types of ZSL methods, i.e. embedding methods, and generative methods.
Embedding methods learn a projection from single modal features to another modal space for similarity measurement~\cite{frome2013devise,lampert2014attribute,li2018discriminative}.
In contrast, generative methods focus on  learning realistic unseen visual distributions from semantic features.
They take advantages of the expressive power of generative adversarial networks (GANs)~\cite{goodfellow2014generative,ye2019unsupervised} to generate plausible visual features for unseen classes~\cite{zhu2018a, xian2018feature, ye2019srgan, rafael2018multi, huang2019generative}.
In this way,  ZSL can be converted to a conventional classification problem.

The connection between semantic and visual relationships is the key of the most ZSL/GZSL methods.
Recently, researchers focus on how to define manually the constraints about the connection.
For example, LsrGAN~\cite{vyas2020leveraging} claims that synthesized visual features of different classes should have a similar relationship to their semantic features.
Thus, they propose to utilize the semantic relationship for guiding visual feature synthesis.
However, semantic features might also be too ambiguous to be classified.
Our previous work, SRGAN~\cite{ye2019srgan} investigates  the visual relationships of different classes and argues that it could be used to rectify their over-smoothing semantics of some classes, and then, synthesize visual features from rectified semantic features.
Other researchers also construct a re-representation space to align visual and semantic features simultaneously~\cite{jiang2018learning}. These methods focus on using single modal information,  semantics or vision. Obviously, single modal information could be incomplete for classification.
Using one modality to constrain the other is imperfect and would cause semantic-to-visual confusion.
Thus, \textit{we pay our attention to two important questions: (1) how to find  more disentangled/separable class representations by utilizing both semantic and visual information? (2) and consequently how to make the best use of disentangled representations for visual feature generation and recognition.}
Our idea is illustrated in Fig.~\ref{fig:approach}.

To answer the first question, we notice that recent studies have designed a series of methods for automatically discriminative representation search~\cite{ye2019srgan,song2018selective,vyas2020leveraging,jiang2018learning,li2018discriminative} where
the  triplet loss is often used~\cite{weinberger2005distance,chen2021entropy,keshari2020generalized,huang2019generative,li2018discriminative}.
For example, Latent Discriminative Features Learning (LDF)~\cite{li2018discriminative} recognizes unseen samples in semantic and latent semantic space, which is searched by triplet loss (TL). 
TL is usually considered in the same fashion among these methods, i.e. they train a metric network (MN) and search a representation space from seen visual features by regulating both inter-class and intra-class distances.
A well-designed MN would minimize the margin among intra-class samples and maximize the margin among inter-class samples.
As a hypothetical result, these works suggest that the unseen classes could also form disentangled searched representations in the searched space.

In this paper, however, we find that TL may lead to a serious problem in ZSL/GZSL due to the inherited nature of ZSL, i.e. the unavailability of unseen visual features.
Particularly,  both visual feature extraction models and MN  cannot access unseen features in ZSL.
Since the feature extraction models are not trained for unseen classes, extracted visual distributions of different unseen classes would be overlapped, leading that unseen features will be  entangled.
Even if MN could search discriminative seen class representations from seen visual features well, the TL training may be highly fragile to out-of-training-distribution features, which is similar to other margin-based losses~\cite{yuan2021in,lyu2021multi}.
As a consequence, MN would produce non-separable and entangled unseen representations due to the under-fitting of unseen visual features, as shown in Fig.~\ref{fig_banner} (a).
As such, we entitle the problem as \emph{entangled unseen visual features problem}.
This problem prevents the ZSL models from achieving the original purpose of using triplet loss, i.e. minimizing the distance among samples of the same classes and maximizing the distance among samples of different classes.


In this work, we propose to address the entangled unseen visual features problem by mitigating the entangled input condition.
To this end, we develop the novel multi-modal triplet loss (MMTL), which combines two modal features, visual and semantic, to form more complete class descriptions.
Compared to the traditional TL, our MMTL can utilize multi-modal information which can benefit disentanglement of the feature representations.
Concretely, when visual features of samples from different classes are close, MMTL can utilize semantic information of samples to distinguish samples, and vice versa.
Therefore, as shown in Fig.~\ref{fig_banner} (b), our MMTL is capable of  searching disentangled representations for all the classes, even when some unseen visual features of different classes are close. In  the testing stage for unseen samples, we can take  sampling methods to obtain unseen representations.
Note that, due to the instability in training  GAN and triplet-based loss, we train our MN and our generator separately in this work.

To answer the second question, we further design the Disentangled Class Representation Generative Adversarial Network (DCRGAN), trying to make the best use of searched representations.
DCR-GAN integrates searched representations in \emph{all} the stages of the ZSL pipeline, i.e. features synthesis, model training, and final classification.
First of all, our generator synthesizes visual features from semantics and searched representations.
Next, for model training, we point out that general GAN-based ZSL adversarial loss is not applicative for ZSL since they adopt a classification loss to make synthesized visual features more discriminative.
However, though such classification loss intends to make synthesized features more separable, our investigation indicates that they cause the real features mixed-up together~\cite{keshari2020generalized}; 
such learned synthesized features would lead to serious misclassification of real samples which are located in class boundaries.
To tackle this problem, we propose our adversarial loss $\mathcal{L}_{WGAN-SR}$ by integrating auxiliary information, i.e. semantic features and searched representations, into our critical loss instead of attaching a classification loss.
In the final classification stage, we train three softmax classifiers in visual, semantic, and searched representations spaces, respectively. Our results show that such integration can largely improve the accuracy both in seen and unseen classes.

Overall, this work has three main contributions.
\begin{itemize}
	\item[1.] We argue that the traditional TL has the inherited shortcoming for ZSL called  the \emph{entangled unseen visual feature problem}, i.e., the traditional TL cannot search appropriately disentangled representations for unseen classes.
	\item[2.] We propose the MMTL to mitigate the entangled unseen feature problem.
	MMTL could search more disentangled representations than the traditional TL, which can be utilized to generate a more realistic distribution.
	\item[3.] We propose a novel GAN-based framework named Disentangled Class Representation Generative Adversarial Network (DCR-GAN) for ZSL.
	DCR-GAN is capable of  searching disentangled representations that are readily integrated in \emph{all} the parts of ZSL. DCR-GAN achieves not only a high accuracy for classifying unseen images but also leads to significant improvement for classifying seen images.
\end{itemize}

\section{Related Work}

\subsection{Zero-shot Learning}
Zero-shot Learning (ZSL)~\cite{xian2017zero,ding2019generative,kodirov2017semantic,zhang2016zero,akata2015evaluation,yang2016zero,ye2019srgan,xian2018feature,qi2017joint,shu2015weakly} is one active research topic in multimedia, which aims at recognizing images from categories that are not included in the training set.
Generalized Zero-Shot Learning proposed in~\cite{chao2016an} considers a more practical situation, in which both seen and unseen instances are mixed in the test data.
One main challenge of ZSL/GZSL is that empirical risk minimization becomes unreliable~\cite{wang2020generalizing}, since unseen visual samples are not available in the training stage.
This challenge also occurs in other relevant problems, i.e., Few-Shot Learning~\cite{zhu2020attribute}.
To overcome the limitations, researchers utilize semantics as intermediate representations of unseen classes.
Such semantics are often manually defined attributes~\cite{lampert2014attribute}, word vectors~\cite{mikolov2013distributed} and text descriptions~\cite{zhu2018a}.
Other works also utilize gaze embedding, that is collected  by non-experts, as semantics~\cite{karessli2017gaze,liu2021goaloriented}.

Mainstreams ZSL methods can fall into embedding methods and generative methods.
Embedding ZSL methods learn a visual-to-semantic embedding~\cite{frome2013devise,lampert2014attribute,li2018discriminative},  a semantic-to-visual embedding~\cite{shigeto2015ridge}, or an unified embedding space~\cite{zhang2015zero,jiang2018learning}.
Generative ZSL methods focus on leveraging Generative Adversarial Network~\cite{goodfellow2014generative} and/or Variational Autoencoders (VAE)~\cite{kingma2013auto} to synthesize unseen visual features from semantic features.

Obviously, the quality of semantic features are the key of all the ZSL methods.
Incomplete semantics would cause confusions of visual features generation.
Semantic Rectifying GAN (SRGAN)~\cite{ye2019srgan} utilizes manually designed distance functions to rectify over-smoothing semantic features by visual similarities.
Some embedding methods~\cite{li2018discriminative} and VAE-based methods~\cite{chen2021entropy,keshari2020generalized} try to utilize the triplet loss to  search automatically more discriminative representations from visual features.

Though previous embedding methods and VAE-based methods have introduced TL to augment class representations, GAN-based methods hardly take concentration on utilizing representations searched by TL or other metric learning (partially due to  their notorious training instability).

In this work, we make an attempt to fill the gap.
We focus on searching automatically  disentangled representations to enhance the fidelity of synthesized features; this is significantly different from the previous ways that manually define constraints between visual and semantic spaces.
We also identify a novel problem of the traditional TL,  and design a framework to utilize the searched representation in training, synthesizing, and recognition stages.

\subsection{Triplet Loss in ZSL}
The traditional TL was  discussed by Google in FaceNet to search face representations for recognition~\cite{schroff2015facenet}.
It takes a metric network (MN) to project an anchor feature $a$, a positive feature $p$, and a negative feature $n$ into the searched representation space.
The anchor and positive features share the same class, while the anchor and negative features  belong to different classes.
MN aims to tighten up the margin between positive pairs $(a,p)$, and widen the margin between negative pairs $(a,n)$.

In previous ZSL works using TL, their MN are all trained by single-modal visual features. For example, one embedding ZSL method,
Latent Discriminative Features Learning (LDF)~\cite{li2018discriminative},  utilizes TL to mine new latent semantic features from visual features.
In generative methods, Entropy-based Uncertainty calibration VAE (EUC-VAE)~\cite{chen2021entropy} and Over-Complete Distribution VAE (OCD-VAE)~\cite{keshari2020generalized} integrate TL in VAE to enhance the separability of encoded representations.
EUC-VAE designs two TLs trained by visual features and semantic features, respectively.
OCD-VAE develops an online batch TL to speed up the process of gradient backward, but it still adopts the same TL formulation as that of LDF.

The above mentioned methods  all ignore the entangled unseen visual features problem.
Traditional TL is highly fragile for out-of-training-distribution features, which is similar to other margin-based losses~\cite{yuan2021in}.
In ZSL, TL is required to search representations not only for seen classes, but also for unseen features.
However, unseen visual features are entangled. Traditional TL in ZSL lacks the ability in defense of overlaps among unseen distributions.
Differently, in this paper, we develop the Multi-Modal Triplet Loss (MMTL) that can mitigate the entangled problem by concatenating multi-modal features to form more complete descriptions of unseen classes.
Benefiting from other modal information, the  unseen distributions do not overlap.
Consequently, our MN trained by MMTL can search disentangled representations which are usually entangled in the traditional TL. As such, MMTL can better meet the intention of using margin-based losses, i.e. maximizing inter-class variation and minimizing intra-class variation.

\section{Disentangled Class Representation Generative Adversarial Network}

The training pipeline of our model follows  two steps:
\begin{enumerate}
    \item[(1)] Pre-training Metric Network (MN, or in short $M$) for searching disentangled representations, and Semantic Rectify Network (SRN or in short $R$) for sampling searched representations from the semantic space. \\This step will be described in Section~\ref{sec:mmtl} and be summarized in Algorithm~\ref{alg:MN}.
 \item[(2)] Training a visual feature generator $G$ with a discriminator $D$ to synthesize pseudo visual features. We also utilize two regressors $F_{1}$ and $F_{2}$ to enhance the multi-modal consistencies among visual space, semantic space, and searched representation space. \\This step will be introduced in Section~\ref{sec:generator} and~\ref{sec:regressor}, and be summarized in Algorithm~\ref{alg:DCRGAN}.
   
\end{enumerate}

Once $G$ is trained , we can train the final ZSL classifier with synthesized unseen features.
Previous generative ZSL methods only train a visual ZSL classifier. Differently in this work, we also train a semantic and a searched representation classifier to make the best of auxiliary information.
The test step will be presented  in Section~\ref{sec:test}.

\subsection{Notations}
Given an image $I$, the proposed model can recognize it as a specific class $c$ even if it is unseen during training.
We take instance $\{\mathbf{x}, \mathbf{a}^{s}, c^{s}\}$ as input in the training stage, where $\mathbf{x}$ describes the instance-level visual feature in the visual feature space $\mathcal{V}$, $\mathbf{a}^{s}$ in the seen semantic space, $\mathcal{A}^{s}$ is class-level semantic extracted from attributes or other description information, and $c^{s}$ denotes the corresponding seen class label.
$\mathcal{C}^{s}$ is the set of seen class labels.
In the testing stage, given an image, ZSL and GZSL will recognize it as an unseen class $\mathcal{C}^{u}$ or a class $c^{s+u}$ (either seen or unseen).
The unseen semantic space and the whole semantic space are denoted as $\mathcal{A}^{u}$, and  $\mathcal{A} = \mathcal{A}^{s} \cup \mathcal{A}^{u}$ respectively.

\begin{figure*}[htbp]
	\centering
	\includegraphics[width=0.8\linewidth]{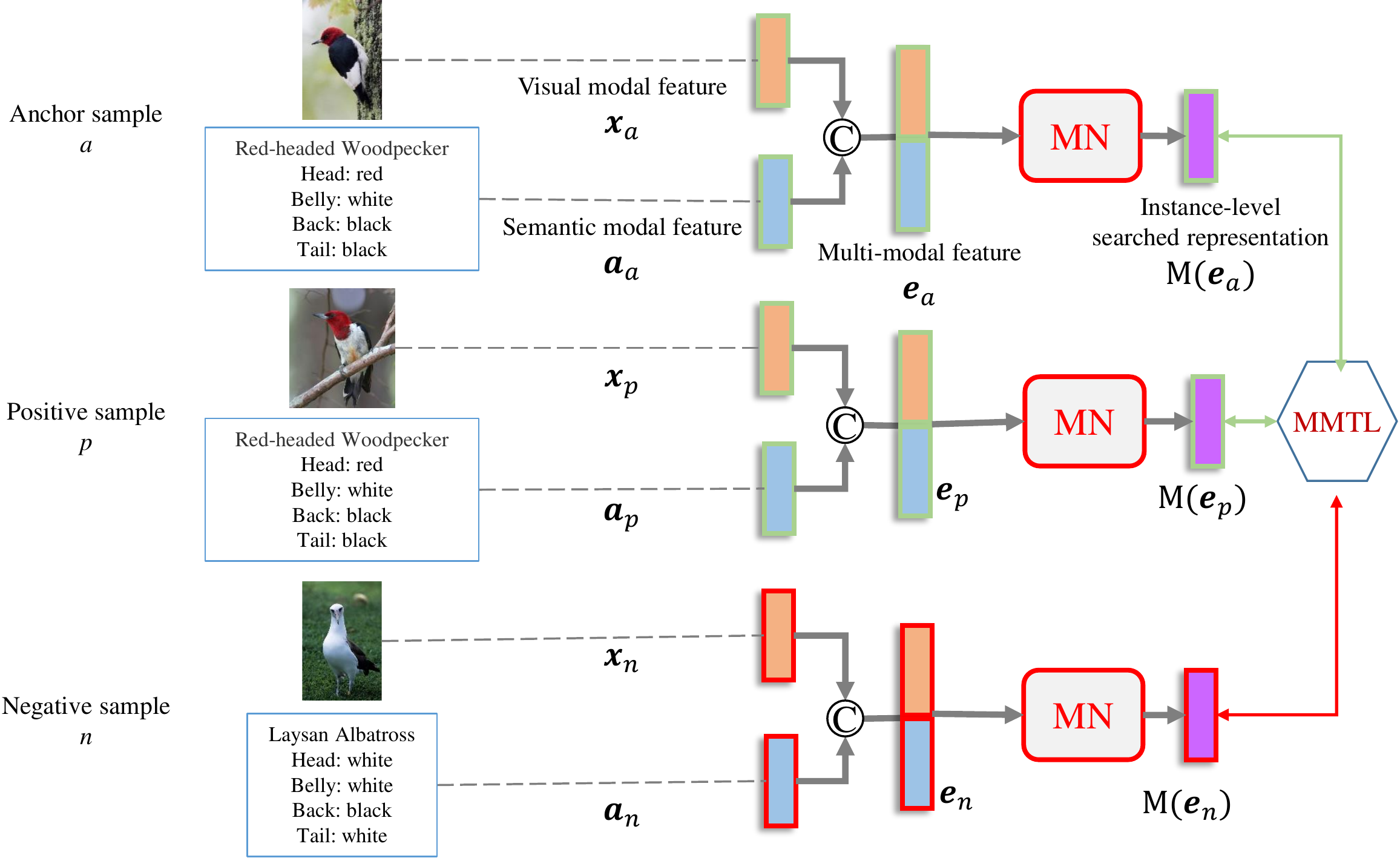}
	\caption{Illustration of our proposed Multi-Modal Triplet Loss (MMTL), which consists of one positive pair ($\mathbf{e}_{a}$,$\mathbf{e}_{p}$) and one negative sample pair ($\mathbf{e}_{a}$,$\mathbf{e}_{n}$).
	Metric Network (MN) trained by our proposed MMTL would minimize the distance of all positive pairs $(M(\mathbf{e}_{a}),M(\mathbf{e}_{p{\tiny }}))$ and maximize the distance of all negative pairs $(M(\mathbf{e}_{a}),M(\mathbf{e}_{n}))$.
	The MN is instance-level.
	}
	\label{fig:MMTL}
\end{figure*}

\begin{figure}[htbp]
	\centering
	\includegraphics[width=0.8\linewidth]{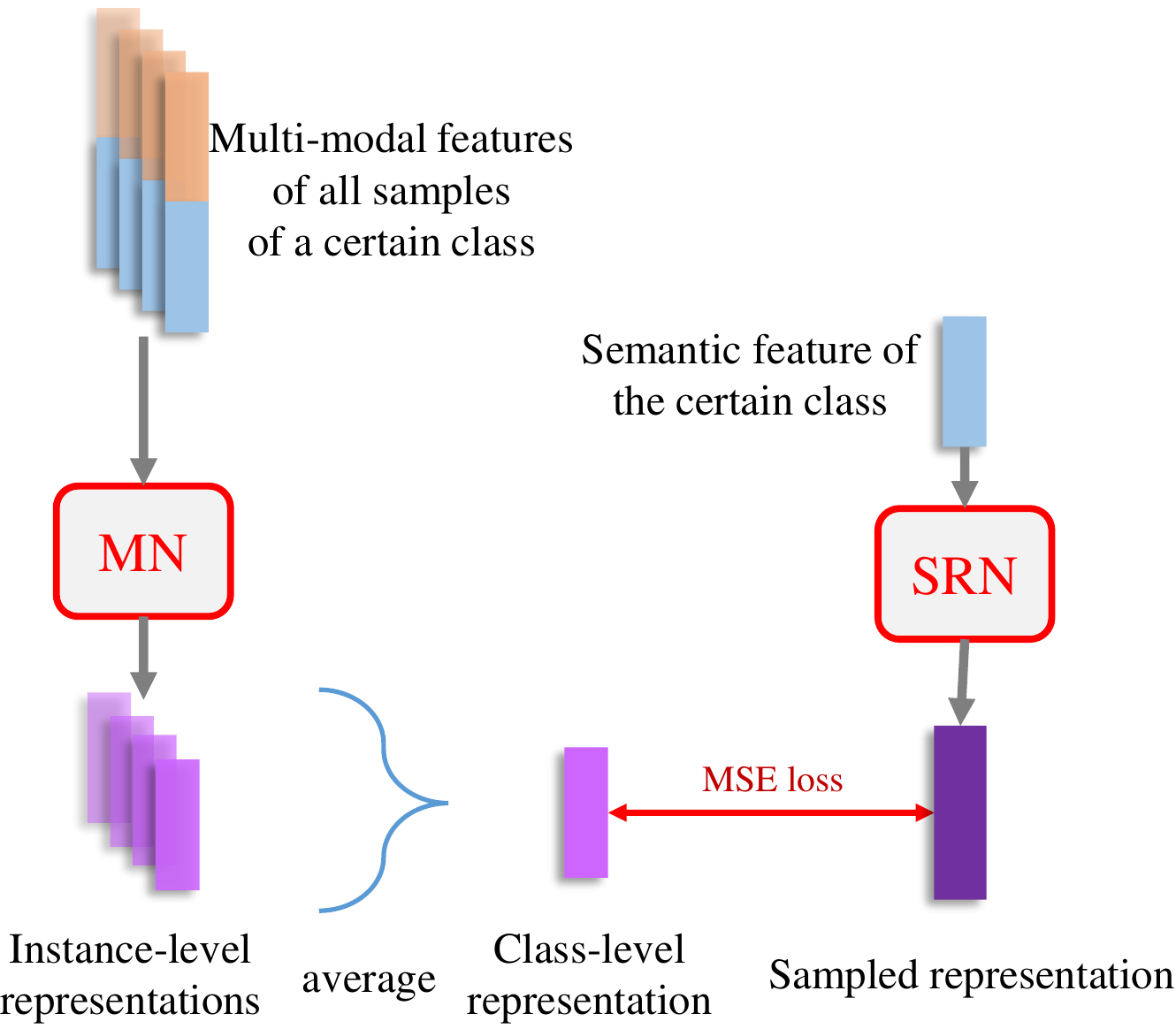}
	\caption{Illustration of the sampling strategy for unseen representations.
		We need  train another network Semantic Rectifying Network (SRN).
		The class-level searched representation is an average of all instance-level representations searched by MN.
		Then, we can train our SRN to learn the mapping: \text{\emph{semantics $\to$ searched representations}}.
		After that, we can obtain unseen searched representations that only need take unseen semantic features as input of SRN.
	}
	\label{fig:SRN}
\end{figure}

\begin{algorithm}[!h]
	\caption{Training algorithm of MN and SRN.}
	\label{alg:MN}
	\begin{algorithmic}[1]
		\REQUIRE The batch size, $m_1$; the number of epoch of training MN, $n_{M}$; the number of epoch of training SRN, $n_{R}$; initial MN parameters, $\theta_{M}$; the margin parameter, $m$; initial SRN parameters, $\theta_{R}$; Adam hyper-parameters, $\alpha$, $\beta_{1}$, and $\beta_{2}$.
		\FOR{$\text{iter}  = 1, \cdots, n_{M}$}
		\STATE Sample a minibatch of seen visual features $\mathbf{x}^{s}$, matching semantic features $\mathbf{a}^{s}$.
		\STATE Compute the MMTL loss $\mathcal{L}_{MMTL}$ using Eq.~\ref{eq:MN}.
		\STATE $\theta_{M} \leftarrow \text{Adam}(\nabla_{\mathcal{L}_{MMTL}},\theta_{M},\alpha,\beta_{1},\beta_{2})$
		\ENDFOR
		\FOR{$\text{iter} = 1, \cdots, n_{R}$}
		\FOR{$t = 1, \cdots, ||\mathcal{C}^{s}||$}
		\STATE Sample all seen visual features $\mathbf{x}^{s}$, and matching semantic features $\mathbf{a}^{s}$ of a certain class in $\mathcal{C}^{s}$.
		\STATE Compute the sampling loss $\mathcal{L}_{sam}$ using Eq.~\ref{eq:sampling}.
		\STATE $\theta_{R} \leftarrow \text{Adam}(\nabla_{\mathcal{L}_{sam}},\theta_{R},\alpha,\beta_{1},\beta_{2})$
		\ENDFOR
		\ENDFOR
		\STATE fix $\theta_{M}$ and $\theta_{R}$
	\end{algorithmic}
\end{algorithm}

\begin{algorithm}[!h]
	\caption{Training algorithm of feature generator.}
	\label{alg:DCRGAN}
	\begin{algorithmic}[1]
		\REQUIRE The maximal loops $N_{loop}$; the batch size $m$; the iteration number of discriminator in a loop $N_d$; the iteration number of generator $N_g$; initial generator parameters $\theta_{G}$; initial discriminator parameters $\theta_{D}$; the trained semantic rectifying Network $R$; the gradient penalty hyper-parameter $\lambda$; the two reconstruction parameters $\lambda_{1}$ and $\lambda_{2}$; Adam hyper-parameters $\alpha$, $\beta_{1}$, and $\beta_{2}$.
		\FOR{$\text{iter} = 1,\cdots,N_{loop}$}
		
		\FOR{$\text{iter} = 1,\cdots,N_{d}$}
		\STATE Sample a mini-batch of seen visual features $\mathbf{x}^{s}$, matching semantic features $\mathbf{a}^{s}$, and random noise $\mathbf{z}$
		\STATE Compute the discriminator loss $\mathcal{L}_{D}$ using Eq.~\ref{eq:DCRGAN}.
		\STATE $\theta_{D} \leftarrow \text{Adam}(\nabla_{\mathcal{L}_{D}},\theta_{D},\alpha,\beta_{1},\beta_{2})$
		\ENDFOR
		
		\FOR{$\text{iter} = 1,\cdots,N_{g}$}
		\STATE Sample a mini-batch of seen visual features $\mathbf{x}^{s}$, corresponding semantic features $\mathbf{a}^{s}$, and random noise $\mathbf{z}$
		\STATE Compute the reconstruction loss $\mathcal{L}_{F_{1}}$ using Eq.~\ref{eq:f1}.
		\STATE $\theta_{F_{1}} \leftarrow \text{Adam}(\nabla_{\mathcal{L}_{F_{1}}},\theta_{F_{1}},\alpha,\beta_{1},\beta_{2})$
		\STATE Compute the reconstruction loss $\mathcal{L}_{F_{2}}$ using Eq.~\ref{eq:f2}.
		\STATE $\theta_{F_{2}} \leftarrow \text{Adam}(\nabla_{\mathcal{L}_{F_{2}}},\theta_{F_{2}},\alpha,\beta_{1},\beta_{2})$
		\STATE Compute the generator loss $\mathcal{L}_{G}$ using Eq.~\ref{eq:DCRGAN}.
		\STATE $\theta_{G} \leftarrow \text{Adam}(\nabla_{\mathcal{L}_{G}},\theta_{G},\alpha,\beta_{1},\beta_{2})$
		\ENDFOR
		
		\ENDFOR
	\end{algorithmic}
\end{algorithm}

\subsection{Multi-modal Triplet Loss and Sampling Strategy}
\label{sec:mmtl}
\subsubsection{Triplet loss in ZSL}

One primary obstacle of generative methods for ZSL mainly comes from the incomplete class semantic features.
Such semantic features would confuse the model, as well as generating less reliable visual features.
To search more comprehensive class representations, previous works try to engage TL to search discriminative class representations.
Given an anchor visual feature $\mathbf{x}^{s}_{a}$ with its label $c^{s}_{a}$, the traditional TL in ZSL attempts to find a positive visual feature $\mathbf{x}^{s}_{p}$ with the same label $c^{s}_{p} = c^{s}_{a}$ and a negative visual feature $\mathbf{x}^{s}_{n}$ with $c^{s}_{n} \neq c^{s}_{a}$.
Then, TL trains a MN by minimizing the distance of the positive pair $d(M(\mathbf{x}^{s}_{a}),M(\mathbf{x}^{s}_{p}))$ and maximizing the distance of the negative pair $d(M(\mathbf{x}^{s}_{a}),M(\mathbf{x}^{s}_{n}))$ by a margin $m$.
MN would search a vector according to the input: $M(\mathbf{x}) \in \mathcal{R}^L$, where $L$ is a hype-parameter to control the size of searched vector.
$d(\cdot,\cdot)$ is usually the Euclidean distance.
It can be formulated as:
\begin{equation}
\mathcal{L}_{TL} = \max(0,m+d(M(\mathbf{x}^{s}_{a}),M(\mathbf{x}^{s}_{p}))-d(M(\mathbf{x}^{s}_{a}),M(\mathbf{x}^{s}_{n}))).
\label{eq:TL}
\end{equation}
The searched representation $M(\mathbf{x}^{s})$ is instance-level representation.
By averaging on all samples of the same classe, we can obtain class-level representations.

\subsubsection{Limitations}
The tradition TL does not consider the \textit{entangled unseen visual features problem} caused by overlapped unseen visual distributions in ZSL.
MN cannot access any unseen features during the  training stage.
For example, there exist some very similar unseen visual features implying $\mathbf{x}^{u}_{i} \approx \mathbf{x}^{u}_{j}$, where $i$ and $j$ belong to two different classes.
Although their semantic features, denoted as $\mathbf{a}^{u}_{i}$ and $\mathbf{a}^{u}_{j}$, are different, visual features $\mathbf{x}^{u}_{i}$ and $\mathbf{x}^{u}_{j}$ are too smoothing to be distinguished by MN.
Such entangled unseen visual features would confuse MN, and finally undermine the discrimivativeness of searched unseen representations, i.e. $M(\mathbf{x}^{u}_{i}) \approx M(\mathbf{x}^{u}_{j})$.

\subsubsection{Multi-modal Triplet Loss}
To tackle the entangled problem, we try to combine two modal features, i.e. both visual and semantic features, to form more complete class descriptions, as shown in Fig.~\ref{fig:MMTL}.
Mathematically, our proposed multi-modal triplet loss can be represented as:
\begin{equation}
\mathcal{L}_{MMTL} = \max(0,m+d(M(\mathbf{e}^{s}_{a}), M(\mathbf{e}^{s}_{p}))-d(M(\mathbf{e}^{s}_{a}), M(\mathbf{e}^{s}_{n}))),
\label{eq:MMTL}
\end{equation}
where $\mathbf{e}$ is a concatenated feature from multiple modalities, e.g. vision, semantics, and/or gaze embedding.
In this work, we concatenate visual and semantic modalities, i.e. $\mathbf{e}^{s} = [\mathbf{x}^{s},\mathbf{a}^{s}]$ where $[\cdot,\cdot]$ denotes the concatenation operation.

Compared to the traditional TL, our MN can utilize multi-modal information to search a latent space which is sharp enough to distinguish different unseen classes.
Obviously, when visual features of samples from different classes are close, MN can utilize semantic information of samples to distinguish samples, and vice versa.
Additionally, we also use the weight decay to prevent over-fitting.
The total loss of training MN is:
\begin{equation}
\mathcal{L}_{MN} = \mathcal{L}_{MMTL} + \|\theta_{M}\|_{2}.
\label{eq:MN}
\end{equation}
Then, we can use sampling methods to get unseen representations from MN for generator training.

\subsubsection{Sampling Unseen Representation Strategy}
In  recognizing unseen samples, we cannot know their labels  before recognition.
However, for a certain unseen class that has $N_{c}$ samples, we need to recognize all visual features of the class to produce its class-level representation $\frac{1}{N_{c}}\sum_{i=1}^{N_{c}}M(\mathbf{e}^{u}_{i})$.
It is completely reversed with the training process of MN, whether using TL or MMTL.
Thus, we need to design a sampling representation strategy to get unseen representations.

The traditional  LDF~\cite{li2018discriminative} method designs a sampling method  by training a relationship matrix $\mathcal{W}$ that maps all seen semantics $\mathcal{A}^{s}$ to all unseen semantics $\mathcal{A}^{s}$, i.e. $\mathcal{A}^{u} \approx \mathcal{W} \cdot \mathcal{A}^{s}$.
Then, unseen representations $M(\mathbf{a}^{u})$ can be obtained from the matrix and searched seen representations, i.e. $M(\mathcal{A}^{u}) = \mathcal{W} \cdot M(\mathcal{A}^{s})$.
Such sampling strategy has several drawbacks: (1) It is a transductive method since it takes unseen semantics to train. (2) It may bring incorrect semantic relationships into the searched representation space.
For example, semantics of some classes are too smoothing to be distinguished~\cite{ye2019srgan}. (3) It forces that the searched representations have the same dimension with semantic features.

In contrast, instead of learning the whole seen-unseen relationships from semantics, we train another network, Semantic Rectifying Network (SRN), for directly mapping semantics, no matter whether seen or unseen, to the searched representation space, as shown in Fig.~\ref{fig:SRN}.
Our sampling strategy  need not unseen semantics in training. Thus it is inductive.
Moreover, it can determine flexibly  the dimensions of the searched representations.
Our experiments also verify that the dimensions of searched representations can affect the ZSL classification.

Specifically, MN will be fixed after we finish its training.
For any seen class, we minimize the $l_2$ loss between the average searched class representation and rectified semantic feature by SRN with the weight decay, i.e.
\begin{equation}
\mathcal{L}_{sam} = \|\frac{1}{N_{c}}\sum_{i=1}^{N_{c}}M(\mathbf{e}^{s}_{i}) - R(\mathbf{a}^{s}_{i})\|_{2} + \|\theta_{R}\|_{2},
\label{eq:sampling}
\end{equation}
where $N_{c}$ is the number of all samples of the chosen seen class.
Then, for an unseen class $u$, we can directly get its searched class representation by rectifying its semantic feature, i.e. $R(\mathbf{a}^u_i)$.
SRN and MN consist of a multi-layer perceptron (MLP) activated by Leaky ReLU, and the output layer does not apply any activation.

\begin{figure}[htbp]
	\centering
	\includegraphics[width=\linewidth]{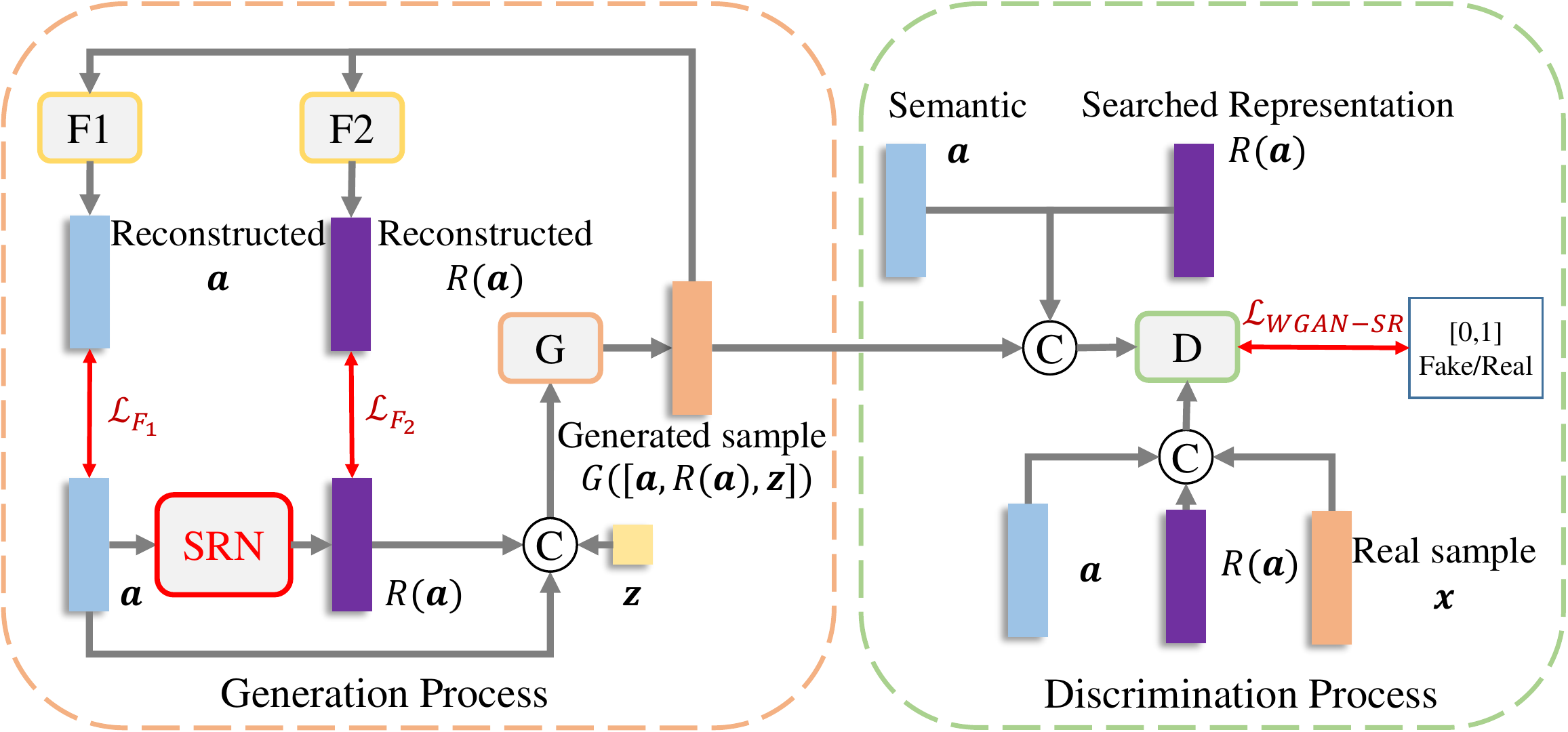}
	\caption{Detailed illustration of the training process for DCR-GAN.
		$G$ and $D$ represent respectively the generator/discriminator of our feature GAN.
		The $SRN$ projects semantic space to searched representation space.
		The $F1$ and $F2$ are two regress networks that respectively projects fake visual features $G([\mathbf{a},R(\mathbf{a}),\mathbf{z}])$ to semantic space and searched class representation space for the inter-class diversity.
	}
	\label{fig:trainGAN}
\end{figure}

\subsection{Visual Feature Synthesis with Search Representation}
\label{sec:generator}

Generative Adversarial Network has demonstrated its usefulness for ZSL~\cite{zhu2018a, xian2018feature, ye2019srgan, rafael2018multi, huang2019generative, long2017from}, due to its promising ability to generate visual features from semantic features.
The most popular generative ZSL methods are based on the conditional WGAN architecture with gradient penalty, which consists of a generator $G$, a discriminator $D$, and a classifier.
The generator $G$ synthesizes visual features from semantic features and normal distribution $z\sim \mathcal{N}(0,1)$,
The discriminator $D$ distinguishes the synthesized samples $\mathbf{x}_{fake}$ from real samples $\mathbf{x}$.
The classifier predicts the probabilities of their label, $logP(y|\mathbf{x}_{fake})$ and $logP(y|\mathbf{x})$.
The classifier, $G$ and $D$ are trained at the same time by the following minimax objective:
\begin{align}
\min \limits_{G} \max \limits_{D} \mathcal{L}_{WGAN} &= \mathbf{E}_{\mathbf{z}\sim p_{\mathbf{z}}}\left[D(\mathbf{x}_{fake})\right]- \mathbf{E}_{\mathbf{x}\sim p_{\mathbf{data}}}\left[D(\mathbf{x})\right]\nonumber\\
&+\mathbf{E}[logP(y|\mathbf{x}_{fake})]-\mathbf{E}[logP(y|\mathbf{x})]\\
 & +\lambda(\|\nabla_{\hat{\mathbf{x}}}D(\hat{\mathbf{x}})\|_{2}-1)^2,\nonumber
\label{eq:WGAN}
\end{align}
where $\mathbf{E}(\cdot)$ denotes the expected value, $\mathbf{x}_{fake} = G(\mathbf{a},\mathbf{z})$, $\lambda$ is a parameter, and the last term $\lambda(\|\nabla_{\hat{\mathbf{x}}}D(\hat{\mathbf{x}})\|_{2}-1)^2$ is the gradient penalty to enforce the Lipschitz constraint~\cite{arjovsky2017wasserstein}, in which $\hat{\mathbf{x}} = \mu\mathbf{x}+(1-\mu)\mathbf{x}_{fake}$ with $\mu \sim U(0,1)$.

However, indiscriminately feeding vague semantic features into a generator may undermine the generated visual features.
By a pre-trained SRN model, we can easily obtain more distinguished class representations.
Therefore, we design a feature GAN model that translates these rectified semantic features into visual features.

$\mathcal{L}_{WGAN}$ has two  limitations: 
(1) It does not consider that semantic and visual space are heteroid.
Some information might be missing completely in the other modal space.
(2) Due to the~\text{entangled unseen visual features problem}, many visual features are in the boundary area of other classes.
In order to reduce the classification risk, $G$ only generates samples that are far from the boundaries of classification, and thus, does not generate hard samples.

To address this problem, we propose the WGAN with the searched representations loss $\mathcal{L}_{WGAN-SR}$:
\begin{align}
\mathcal{L}_{WGAN-SR} &= \mathbf{E}_{\mathbf{z}\sim p_{\mathbf{z}}}\left[D([\mathbf{x}_{fake},\mathbf{a},R(\mathbf{a})])\right]\nonumber\\
&- \mathbf{E}_{\mathbf{x}\sim p_{\mathbf{data}}}\left[D([\mathbf{x},\mathbf{a},R(\mathbf{a})])\right]\\
&+\lambda(\|\nabla_{\hat{\mathbf{x}}}D([\hat{\mathbf{x}},\mathbf{a},R(\mathbf{a})])\|_{2}-1)^2\nonumber,
\label{eq:WGANSR}
\end{align}
where $\mathbf{x}_{fake} = G(\mathbf{a},R(\mathbf{a}),\mathbf{z})$.
The training process of our DCR-GAN is described in Fig.~\ref{fig:trainGAN}.
Our $\mathcal{L}_{WGAN-SR}$ enjoys two differences with $\mathcal{L}_{WGAN}$:
(1) We integrate searched representations to align two modalities.
(2) We remove the classifier and leverage auxiliary information, i.e. semantic features and searched representations, to train a class-sensitive discriminator $D$.
With the integrated auxiliary information, interlaced class boundaries are pushed off.  
In this case, our generator $G$ does not worry about the classification risk for hard samples.

\subsection{Feature Reconstruction}
\label{sec:regressor}
With the above process, our model is able to synthesize proper visual features to some extent.
However, there exists a significant problem, e.g. the generated visual features have poor consistencies with the input semantics and searched representations.
Accordingly, we utilize two regression networks to keep the consistencies of \emph{semantic $\to$ visual $\to$ semantic} space and \emph{searched representation $\to$ visual $\to$ searched representation} space, respectively.
Specifically, the regression network $F_{1}$, keeping in step with the generator $G$, takes the generated feature $\mathbf{x}_{fake}$ as input, and builds consistency losses between original semantics and reconstructed semantics from visual features.
The regression network $F_{2}$ works in the same way for searched representations.
The two regression loss for $F_{1}$ and $F_{2}$ can be computed by
\begin{equation}
\mathcal{L}_{F_1} = \|F_{1}(\mathbf{x}_{fake})-\mathbf{a}\|_{1},
\label{eq:f1}
\end{equation}
\begin{equation}
\mathcal{L}_{F_2} = \|F_{2}(\mathbf{x}_{fake})-R(\mathbf{a})\|_{1}.
\label{eq:f2}
\end{equation}
Obviously, $G\circ F_{1}$ and $G\circ F_{2}$ can be considered as two Auto-Encoders~\cite{kingma2013auto}, where $A \circ B$ denotes the composite of two mappings.
The reconstruction $G\circ F_{1}$ enhances the relationship between the synthetic visual features and the corresponding class semantics by minimizing the difference between the reconstructed and original semantic features.
The searched representation reconstruction $G\circ F_{2}$ works in the same way.

Finally, by integrating the reconstruction losses, the new objective of our DCR-GAN can be modified as:
\begin{align}
\min \limits_{G,F_{1},F_{2}} \max \limits_{D} \mathcal{L}_{DCRGAN} = \mathcal{L}_{WGAN-SR} + \lambda_{1} \mathcal{L}_{F1} + \lambda_{2} \mathcal{L}_{F2},
\label{eq:DCRGAN}
\end{align}
where $\lambda_{1}$ and $\lambda_{2}$ are two reconstruction parameters for semantic and searched representation, respectively.
\begin{figure}[htbp]
	\centering
	\includegraphics[width=0.95\linewidth]{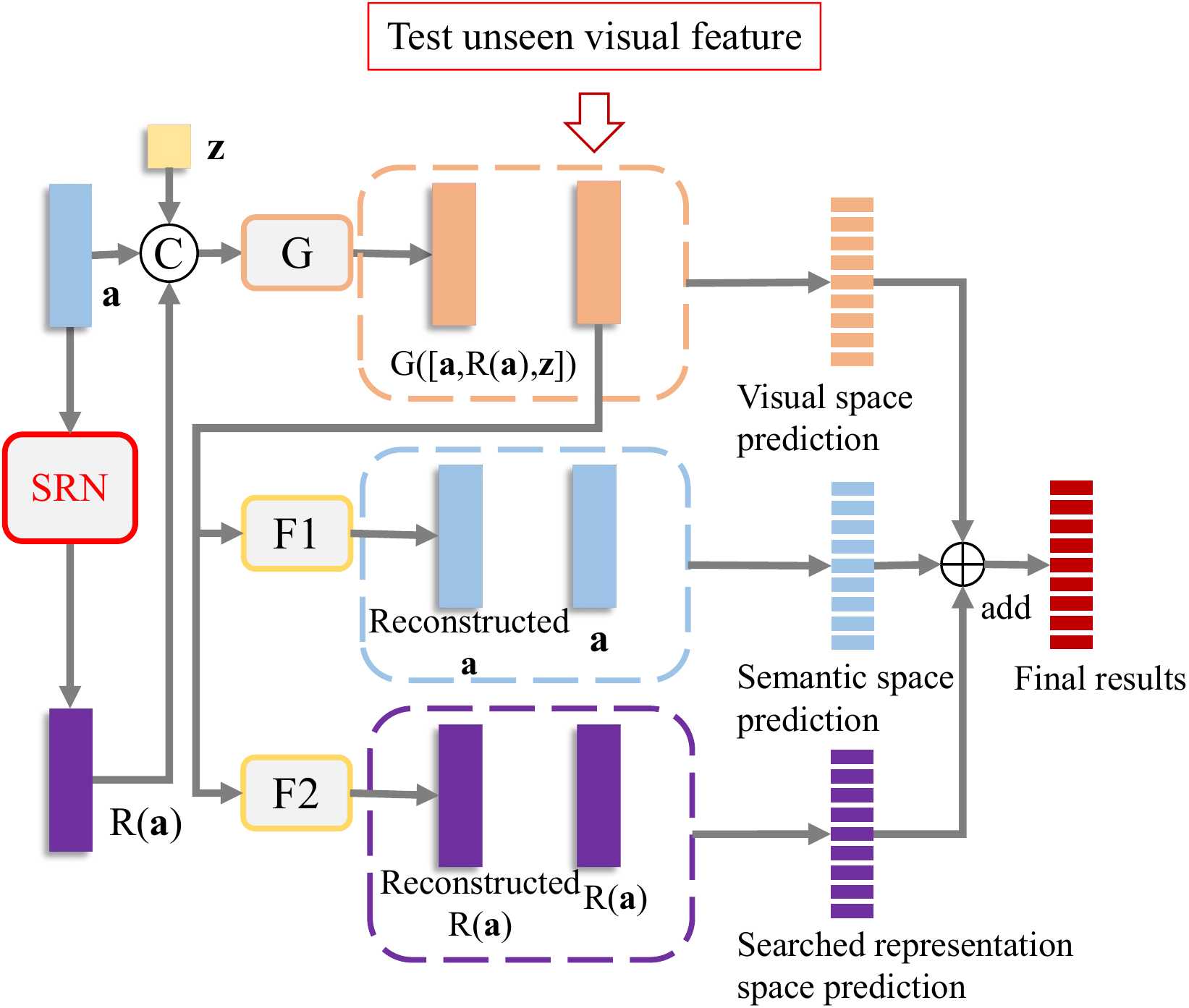}
	\caption{Overview of zero-shot classification.
	We use our trained generator for unseen visual feature synthesis.
	Then, the synthesized feature is used to train a softmax classifier.
	Analogously, a semantic classifier and  searched representation classifier are trained.
	We also use $F_{1}$ and $F_{2}$ to map all the real unseen visual features into the semantic space and searched representation space, respectively.
	We integrate the final results from visual, semantic, and searched representation space.
}
	\label{fig:test}
\end{figure}
\subsection{Zero-shot Classification with Searched Representation}
\label{sec:test}
By Eq.~\ref{eq:DCRGAN}, we can train a GAN generator $G$ which is able to synthesize virtual visual features for unseen categories.
Then, we use the synthesized features to train a classifier, e.g. softmax, to recognize the real unseen instances.
In other words, zero-shot learning is converted into a supervised classification problem that is performed in the visual space.
As shown in Fig.~\ref{fig:test}, we train a softmax classifier by synthesized unseen visual features.
We also use $F_{1}$ and $F_{2}$ to map all the real unseen visual features into the semantic space and searched representation space, respectively.
Analogously, a semantic classifier and  searched representation classifier are trained.
Thus, in our model, we take full advantages of the visual space prediction score $f_{VS}$, semantic space prediction score $f_{SS}$, and searched representation space prediction score $f_{SRS}$.
Finally, we get the final classification scores as:
\begin{equation}
	\label{equation:finalCls}
	f = f_{VS}+ \omega_{1}f_{SS} + \omega_{2}f_{SRS},
\end{equation}
where $\omega_{1}$ and $\omega_{2}$ are two parameters used to balance the three terms, as shown in Fig.~\ref{fig:test}.

For GZSL, the main steps are the same as ZSL.
The only difference lies in the final classification process.
Specifically, the testing data of GZSL are from both the seen and unseen categories.
Thus, we need to train a classifier on both real seen features and synthesized unseen features.

\section{Experiments}

\subsection{Datasets}
\begin{table*}[htbp]
	\centering
	\caption{Statistics of datasets.}
	\label{table:dataset}
	\begin{tabular}{c|c|c|c|c|c|c}
		\hline
		Dataset&$\#$attributes&\tabincell{c}{$\#$seen classes\\(train+val)}&$\#$unseen classes&\tabincell{c}{$\#$images\\(total)}&\tabincell{c}{$\#$images\\(train+val)}&\tabincell{c}{$\#$images\\(test unseen/seen)}\\
		\hline
		AWA1~\cite{lampert2009learning}&85&27+13&10&30475&19832&4958/5685\\
		\hline
		CUB~\cite{wah2011caltech}&312&100+50&50&11788&7057&2679/1764\\
		\hline
		APY~\cite{farhadi2009describing}&64&15+5&12&15339&5932&7924/1483\\
		\hline
		SUN~\cite{patterson2014the}&102&580+65&72&14340&10320&1440/2580\\
		\hline
	\end{tabular}
\end{table*}
We evaluate our approach on four benchmark datasets for ZSL and GZSL: (1) Caltech-UCSD-Birds 200-2011 (CUB) \cite{wah2011caltech} consisting of 11,788 images of 200 classes of birds annotated with 312 binary attributes; (2) Animals with Attributes (AWA) \cite{lampert2009learning} consisting of 30,475 images of 50 animals classes with 85 attributes; (3) Attribute Pascal and Yahoo (APY) \cite{farhadi2009describing} containing of 15,339 images, 32 classes and 64 attributes from both PASCAL VOC 2008 dataset and Yahoo image search engine; (4) SUN Attribute (SUN) \cite{patterson2014the} annotating 102 attributes on 14,340 images from 717 types of scene.
For the four datasets, we use the widely-used ZSL and GZSL split proposed in \cite{xian2017zero}.
For clarity, the statistics of these datasets are summarized in Table~\ref{table:dataset}.

We adopt the evaluation metrics proposed in \cite{xian2017zero}.
For ZSL, we measure \emph{average per-class top-1 accuracy} ($T1$) of unseen classes $C_{u}$.
It is defined as follows:
\begin{equation}
	T1 = \frac{1}{\|\mathcal{C}^{u}\|} \sum_{i}^{\|\mathcal{C}^{u}\|} \frac{\# \text{ of correction predictions in }i}{\# \text{ of samples in i}}.
\end{equation}
For GZSL, we compute the average per-class top-1 accuracy of seen classes $C_{s}$, denoted by $S$, the average per-class top-1 accuracy of unseen classes $C_{u}$, denoted by $U$, and their harmonic mean, i.e. $H = 2 \times (S \times U)/(S + U)$.

\subsection{Comparison to State-of-the-arts}
\begin{table*}[htbp]
	\centering
	\caption{Comparison with the state-of-the-art methods on four datasets.}
	\label{table:OverallComparison}
	\begin{tabular}{p{2.7cm}p{0.2cm}p{0.2cm}p{0.2cm}p{0.2cm}| p{0.3cm}p{0.3cm}p{0.3cm}p{0.3cm}p{0.3cm}p{0.3cm}p{0.3cm}p{0.3cm}p{0.3cm}p{0.3cm}p{0.3cm}p{0.3cm}}
		\hline
		& \multicolumn{4}{|c|}{Zero-shot Learning} & \multicolumn{12}{|c}{Generalizaed Zero-shot Learning}\\
		\hline
		& \multicolumn{1}{|c}{AWA1} & \multicolumn{1}{c}{CUB} & \multicolumn{1}{c}{APY} & \multicolumn{1}{c}{SUN} & \multicolumn{3}{|c|}{AWA1} & \multicolumn{3}{|c|}{CUB} &\multicolumn{3}{|c|}{APY} &\multicolumn{3}{|c}{SUN} \\
		\hline
		Approach & \multicolumn{1}{|c}{T1} & \multicolumn{1}{c}{T1} & \multicolumn{1}{c}{T1} & \multicolumn{1}{c|}{T1} & \multicolumn{1}{|c}{U} & \multicolumn{1}{c}{S} & \multicolumn{1}{c|}{H} & \multicolumn{1}{|c}{U} & \multicolumn{1}{c}{S} & \multicolumn{1}{c|}{H} & \multicolumn{1}{|c}{U} & \multicolumn{1}{c}{S} & \multicolumn{1}{c|}{H} & \multicolumn{1}{|c}{U} & \multicolumn{1}{c}{S} & \multicolumn{1}{c}{H} \\
		\hline
		\multicolumn{17}{c}{Embedding approaches}\\
		\hline
		DEVISE~\cite{frome2013devise} & \multicolumn{1}{|c}{54.2} & \multicolumn{1}{c}{52.0} & \multicolumn{1}{c}{39.8} & \multicolumn{1}{c|}{56.5} & \multicolumn{1}{|c}{13.4} & \multicolumn{1}{c}{68.7} & \multicolumn{1}{c|}{22.4} & \multicolumn{1}{|c}{23.8} & \multicolumn{1}{c}{53.0} & \multicolumn{1}{c|}{32.8} & \multicolumn{1}{|c}{4.9} & \multicolumn{1}{c}{76.9} & \multicolumn{1}{c|}{9.2} & \multicolumn{1}{|c}{16.9} & \multicolumn{1}{c}{27.4} & \multicolumn{1}{c}{20.9} \\
		DAP~\cite{lampert2014attribute} & \multicolumn{1}{|c}{44.1} & \multicolumn{1}{c}{40.0} & \multicolumn{1}{c}{33.8} & \multicolumn{1}{c|}{39.9} & \multicolumn{1}{|c}{0.0} & \multicolumn{1}{c}{\textbf{88.7}} & \multicolumn{1}{c|}{0.0} & \multicolumn{1}{|c}{1.7} & \multicolumn{1}{c}{67.9} & \multicolumn{1}{c|}{3.3} & \multicolumn{1}{|c}{4.8} & \multicolumn{1}{c}{78.3} & \multicolumn{1}{c|}{9.0} & \multicolumn{1}{|c}{4.2} & \multicolumn{1}{c}{25.1} & \multicolumn{1}{c}{7.2}\\
		SSE~\cite{zhang2015zero} & \multicolumn{1}{|c}{60.1} & \multicolumn{1}{c}{43.9} & \multicolumn{1}{c}{34.0} & \multicolumn{1}{c|}{51.5} & \multicolumn{1}{|c}{7.0} & \multicolumn{1}{c}{80.5} & \multicolumn{1}{c|}{12.9} & \multicolumn{1}{|c}{8.5} & \multicolumn{1}{c}{46.9} & \multicolumn{1}{c|}{14.4} & \multicolumn{1}{|c}{0.2} & \multicolumn{1}{c}{78.9} & \multicolumn{1}{c|}{0.4} & \multicolumn{1}{|c}{2.1} & \multicolumn{1}{c}{36.4} & \multicolumn{1}{c}{4.0} \\
		SJE~\cite{akata2015evaluation} & \multicolumn{1}{|c}{65.6} & \multicolumn{1}{c}{53.9} & \multicolumn{1}{c}{32.9} & \multicolumn{1}{c|}{53.7} & \multicolumn{1}{|c}{11.3} & \multicolumn{1}{c}{74.6} & \multicolumn{1}{c|}{19.6} & \multicolumn{1}{|c}{23.5} & \multicolumn{1}{c}{59.2} & \multicolumn{1}{c|}{33.6} & \multicolumn{1}{|c}{3.7} & \multicolumn{1}{c}{55.7} & \multicolumn{1}{c|}{6.9} & \multicolumn{1}{|c}{14.7} & \multicolumn{1}{c}{30.5} & \multicolumn{1}{c}{19.8} \\
		ESZSL~\cite{bernardino2015an} & \multicolumn{1}{|c}{58.2} & \multicolumn{1}{c}{53.9} & \multicolumn{1}{c}{38.3} & \multicolumn{1}{c|}{54.5} & \multicolumn{1}{|c}{6.6} & \multicolumn{1}{c}{75.6} & \multicolumn{1}{c|}{12.1} & \multicolumn{1}{|c}{12.6} & \multicolumn{1}{c}{63.8} & \multicolumn{1}{c|}{21.0} & \multicolumn{1}{|c}{2.4} & \multicolumn{1}{c}{70.1} & \multicolumn{1}{c|}{4.6} & \multicolumn{1}{|c}{11.0} & \multicolumn{1}{c}{27.9} & \multicolumn{1}{c}{15.8} \\
		ALE~\cite{akata2016label} & \multicolumn{1}{|c}{59.9} & \multicolumn{1}{c}{54.9} & \multicolumn{1}{c}{39.7} & \multicolumn{1}{c|}{58.1} & \multicolumn{1}{|c}{16.8} & \multicolumn{1}{c}{76.1} & \multicolumn{1}{c|}{27.5} & \multicolumn{1}{|c}{23.7} & \multicolumn{1}{c}{62.8} & \multicolumn{1}{c|}{34.4} & \multicolumn{1}{|c}{4.6} & \multicolumn{1}{c}{73.7} & \multicolumn{1}{c|}{8.7} & \multicolumn{1}{|c}{21.8} & \multicolumn{1}{c}{33.1} & \multicolumn{1}{c}{26.3} \\
		LATEM~\cite{xian2016latent} & \multicolumn{1}{|c}{55.1} & \multicolumn{1}{c}{49.3} & \multicolumn{1}{c}{35.2} & \multicolumn{1}{c|}{55.3} & \multicolumn{1}{|c}{7.3} & \multicolumn{1}{c}{71.7} & \multicolumn{1}{c|}{13.3} & \multicolumn{1}{|c}{15.2} & \multicolumn{1}{c}{57.3} & \multicolumn{1}{c|}{24.0} & \multicolumn{1}{|c}{0.1} & \multicolumn{1}{c}{73.0} & \multicolumn{1}{c|}{0.2} & \multicolumn{1}{|c}{14.7} & \multicolumn{1}{c}{28.8} & \multicolumn{1}{c}{19.5} \\
		SYNC~\cite{changpinyo2016synthesized} & \multicolumn{1}{|c}{54.0} & \multicolumn{1}{c}{55.6} & \multicolumn{1}{c}{23.9} & \multicolumn{1}{c|}{56.3} & \multicolumn{1}{|c}{8.9} & \multicolumn{1}{c}{87.3} & \multicolumn{1}{c|}{16.2} & \multicolumn{1}{|c}{11.5} & \multicolumn{1}{c}{70.9} & \multicolumn{1}{c|}{19.8} & \multicolumn{1}{|c}{7.4} & \multicolumn{1}{c}{66.3} & \multicolumn{1}{c|}{13.3} & \multicolumn{1}{|c}{7.9} & \multicolumn{1}{c}{43.3} & \multicolumn{1}{c}{13.4} \\
		SAE~\cite{kodirov2017semantic} & \multicolumn{1}{|c}{53.0} & \multicolumn{1}{c}{33.3} & \multicolumn{1}{c}{8.3} & \multicolumn{1}{c|}{40.3} & \multicolumn{1}{|c}{1.8} & \multicolumn{1}{c}{77.1} & \multicolumn{1}{c|}{3.5} & \multicolumn{1}{|c}{7.8} & \multicolumn{1}{c}{54.0} & \multicolumn{1}{c|}{13.6} & \multicolumn{1}{|c}{0.4} & \multicolumn{1}{c}{\textbf{80.9}} & \multicolumn{1}{c|}{0.9} & \multicolumn{1}{|c}{8.8} & \multicolumn{1}{c}{18.0} & \multicolumn{1}{c}{11.8} \\
		PSR~\cite{annadani2018preserving} & \multicolumn{1}{|c}{-} & \multicolumn{1}{c}{56.0} & \multicolumn{1}{c}{38.4} & \multicolumn{1}{c|}{61.4} & \multicolumn{1}{|c}{-} & \multicolumn{1}{c}{-} & \multicolumn{1}{c|}{-} & \multicolumn{1}{|c}{24.6} & \multicolumn{1}{c}{54.3} & \multicolumn{1}{c|}{33.9} & \multicolumn{1}{|c}{13.5} & \multicolumn{1}{c}{51.4} & \multicolumn{1}{c|}{21.4} & \multicolumn{1}{|c}{20.8} & \multicolumn{1}{c}{37.2} & \multicolumn{1}{c}{26.7} \\
		CDL~\cite{jiang2018learning} & \multicolumn{1}{|c}{69.9} & \multicolumn{1}{c}{54.5} & \multicolumn{1}{c}{43.0} & \multicolumn{1}{c|}{63.6} & \multicolumn{1}{|c}{28.1} & \multicolumn{1}{c}{73.5} & \multicolumn{1}{c|}{40.6} & \multicolumn{1}{|c}{23.5} & \multicolumn{1}{c}{55.2} & \multicolumn{1}{c|}{32.9} & \multicolumn{1}{|c}{19.8} & \multicolumn{1}{c}{48.6} & \multicolumn{1}{c|}{28.1} & \multicolumn{1}{|c}{21.5} & \multicolumn{1}{c}{34.7} & \multicolumn{1}{c}{26.5} \\
		CRNet~\cite{zhang2019co} & \multicolumn{1}{|c}{-} & \multicolumn{1}{c}{-}& \multicolumn{1}{c}{-} & \multicolumn{1}{c|}{-} & \multicolumn{1}{c}{58.1} & \multicolumn{1}{c}{74.7} & \multicolumn{1}{c|}{65.4} & \multicolumn{1}{|c}{45.5} & \multicolumn{1}{c}{56.8} & \multicolumn{1}{c|}{50.5} & \multicolumn{1}{|c}{32.4} & \multicolumn{1}{c}{68.4} & \multicolumn{1}{c|}{44.0} & \multicolumn{1}{|c}{34.1} & \multicolumn{1}{c}{36.5} & \multicolumn{1}{c}{35.3}\\
		DVBE (fixing)~\cite{min2020domain} & \multicolumn{1}{|c}{-} & \multicolumn{1}{c}{-}& \multicolumn{1}{c}{-} & \multicolumn{1}{c|}{-} & \multicolumn{1}{|c}{-}& \multicolumn{1}{c}{-} & \multicolumn{1}{c|}{-} & \multicolumn{1}{|c}{53.2} & \multicolumn{1}{c}{60.2} & \multicolumn{1}{c|}{56.5} & \multicolumn{1}{|c}{32.6} & \multicolumn{1}{c}{58.3} & \multicolumn{1}{c|}{41.8} & \multicolumn{1}{|c}{45.0} & \multicolumn{1}{c}{37.2} & \multicolumn{1}{c}{40.7}\\
		\hline
		\multicolumn{17}{c}{Generative approaches}\\
		\hline
		f-CLSWGAN~\cite{xian2018feature} & \multicolumn{1}{|c}{68.2} & \multicolumn{1}{c}{57.3} & \multicolumn{1}{c}{40.5} & \multicolumn{1}{c|}{60.8} &
		\multicolumn{1}{|c}{57.9} & \multicolumn{1}{c}{61.4} & \multicolumn{1}{c|}{59.6} &
		\multicolumn{1}{|c}{43.7} & \multicolumn{1}{c}{57.7} & \multicolumn{1}{c|}{49.7} &
		\multicolumn{1}{|c}{32.9} & \multicolumn{1}{c}{61.7} & \multicolumn{1}{c|}{42.9} &
		\multicolumn{1}{|c}{42.6} & \multicolumn{1}{c}{36.6} & \multicolumn{1}{c}{39.4}\\
		cycle-CLSWGAN~\cite{rafael2018multi} & \multicolumn{1}{|c}{66.3} & \multicolumn{1}{c}{58.6} & \multicolumn{1}{c}{-} & \multicolumn{1}{c|}{59.9} &
		\multicolumn{1}{|c}{56.9} & \multicolumn{1}{c}{64.0} & \multicolumn{1}{c|}{60.2} &
		\multicolumn{1}{|c}{45.7} & \multicolumn{1}{c}{61.0} & \multicolumn{1}{c|}{52.3} &
		\multicolumn{1}{|c}{-} & \multicolumn{1}{c}{-} & \multicolumn{1}{c|}{-} &
		\multicolumn{1}{|c}{\textbf{49.4}} & \multicolumn{1}{c}{33.6} & \multicolumn{1}{c}{40.0}\\
		DASCN~\cite{ni2019dual} & \multicolumn{1}{|c}{-} & \multicolumn{1}{c}{-} & \multicolumn{1}{c}{-} & \multicolumn{1}{c|}{-} & \multicolumn{1}{c}{59.3} & \multicolumn{1}{c}{68.0} & \multicolumn{1}{c|}{63.4} & \multicolumn{1}{|c}{45.9} & \multicolumn{1}{c}{59.0} & \multicolumn{1}{c|}{51.6} & \multicolumn{1}{|c}{\textbf{39.7}} & \multicolumn{1}{c}{59.5} & \multicolumn{1}{c|}{47.6} & \multicolumn{1}{|c}{42.4} & \multicolumn{1}{c}{38.5} & \multicolumn{1}{c}{40.3} \\
		AFC-GAN~\cite{li2019alleviating} & \multicolumn{1}{|c}{69.1} & \multicolumn{1}{c}{\textbf{62.9}} & \multicolumn{1}{c}{45.5} & \multicolumn{1}{c|}{63.3} & \multicolumn{1}{c}{58.2} & \multicolumn{1}{c}{66.8} & \multicolumn{1}{c|}{62.2} & \multicolumn{1}{|c}{53.5} & \multicolumn{1}{c}{59.7} & \multicolumn{1}{c|}{56.4} & \multicolumn{1}{|c}{36.5} & \multicolumn{1}{c}{62.6} & \multicolumn{1}{c|}{46.1} & \multicolumn{1}{|c}{49.1} & \multicolumn{1}{c}{36.1} & \multicolumn{1}{c}{41.6} \\
		GDAN~\cite{huang2019generative} & \multicolumn{1}{|c}{-} & \multicolumn{1}{c}{-} & \multicolumn{1}{c}{-} & \multicolumn{1}{c|}{-} & \multicolumn{1}{|c}{-} & \multicolumn{1}{c}{-} & \multicolumn{1}{c|}{-} & \multicolumn{1}{|c}{39.3} & \multicolumn{1}{c}{66.7} & \multicolumn{1}{c|}{49.5} & \multicolumn{1}{|c}{30.4} & \multicolumn{1}{c}{75.0} & \multicolumn{1}{c|}{43.4} & \multicolumn{1}{|c}{38.1} & \multicolumn{1}{c}{\textbf{89.9}} & \multicolumn{1}{c}{\textbf{53.4}}\\
		GAZSL~\cite{zhu2018a} & \multicolumn{1}{|c}{68.2} & \multicolumn{1}{c}{55.8} & \multicolumn{1}{c}{41.1} & \multicolumn{1}{c|}{61.3} & \multicolumn{1}{|c}{19.2} & \multicolumn{1}{c}{86.5} & \multicolumn{1}{c|}{31.4} & \multicolumn{1}{|c}{23.9} & \multicolumn{1}{c}{60.6} & \multicolumn{1}{c|}{34.3} & \multicolumn{1}{|c}{14.2} & \multicolumn{1}{c}{78.6} & \multicolumn{1}{c|}{24.0} & \multicolumn{1}{|c}{21.7} & \multicolumn{1}{c}{34.5} & \multicolumn{1}{c}{26.7} \\
		SRGAN~\cite{ye2019srgan} & \multicolumn{1}{|c}{\textbf{72.0}} & \multicolumn{1}{c}{55.4} & \multicolumn{1}{c}{44.0} & \multicolumn{1}{c|}{62.3} & \multicolumn{1}{|c}{41.5} & \multicolumn{1}{c}{83.1} & \multicolumn{1}{c|}{55.3} & \multicolumn{1}{|c}{31.3} & \multicolumn{1}{c}{60.9} & \multicolumn{1}{c|}{41.3} & \multicolumn{1}{|c}{22.3} & \multicolumn{1}{c}{78.4} & \multicolumn{1}{c|}{34.8} & \multicolumn{1}{|c}{22.1} & \multicolumn{1}{c}{38.3} & \multicolumn{1}{c}{27.4}\\
		OCD-VAE~\cite{keshari2020generalized} & \multicolumn{1}{|c}{-} & \multicolumn{1}{c}{60.3}& \multicolumn{1}{c}{-} & \multicolumn{1}{c|}{63.5} & \multicolumn{1}{|c}{-}& \multicolumn{1}{c}{-} & \multicolumn{1}{c|}{-} & \multicolumn{1}{|c}{44.8} & \multicolumn{1}{c}{59.9} & \multicolumn{1}{c|}{51.3} & \multicolumn{1}{|c}{-} & \multicolumn{1}{c}{-} & \multicolumn{1}{c|}{-} & \multicolumn{1}{|c}{44.8} & \multicolumn{1}{c}{42.9} & \multicolumn{1}{c}{43.8}\\
		EUC-VAE~\cite{chen2021entropy} & \multicolumn{1}{|c}{65.7} & \multicolumn{1}{c}{61.7}& \multicolumn{1}{c}{39.1} & \multicolumn{1}{c|}{63.5} & \multicolumn{1}{|c}{60.4}& \multicolumn{1}{c}{70.4} & \multicolumn{1}{c|}{65.1} & \multicolumn{1}{|c}{50.8} & \multicolumn{1}{c}{55.1} & \multicolumn{1}{c|}{52.9} & \multicolumn{1}{|c}{44.1} & \multicolumn{1}{c}{36.8} & \multicolumn{1}{c|}{40.1} & \multicolumn{1}{|c}{35.0} & \multicolumn{1}{c}{62.7} & \multicolumn{1}{c}{44.9}\\
		LsrGAN~\cite{vyas2020leveraging} & \multicolumn{1}{|c}{66.4} & \multicolumn{1}{c}{60.3} & \multicolumn{1}{c}{-} & \multicolumn{1}{c|}{62.5} & \multicolumn{1}{|c}{54.6} & \multicolumn{1}{c}{74.6} & \multicolumn{1}{c|}{63.0} & \multicolumn{1}{|c}{48.1} & \multicolumn{1}{c}{59.1} & \multicolumn{1}{c|}{53.0} & \multicolumn{1}{|c}{-} & \multicolumn{1}{c}{-} & \multicolumn{1}{c|}{-} & \multicolumn{1}{|c}{44.8} & \multicolumn{1}{c}{37.7} & \multicolumn{1}{c}{40.9} \\
		
		DCR-GAN (ours) & \multicolumn{1}{|c}{71.0} & \multicolumn{1}{c}{61.0} & \multicolumn{1}{c}{\textbf{48.0}} & \multicolumn{1}{c|}{\textbf{63.7}} & \multicolumn{1}{c}{\textbf{62.7}} & \multicolumn{1}{c}{73.3} & \multicolumn{1}{c|}{\textbf{67.6}} & \multicolumn{1}{|c}{\textbf{55.8}} & \multicolumn{1}{c}{\textbf{66.8}} & \multicolumn{1}{c|}{\textbf{60.8}} & \multicolumn{1}{|c}{37.2} & \multicolumn{1}{c}{71.7} & \multicolumn{1}{c|}{\textbf{49.0}} & \multicolumn{1}{|c}{47.1} & \multicolumn{1}{c}{38.5} & \multicolumn{1}{c}{42.4} \\
		\hline
	\end{tabular}
\end{table*}

\begin{table*}[htbp]
	\centering
	\caption{Ablation study in the  GZSL setting. The results are reported as average per-class top-1 accuracy of unseen classes (U), seen classes (S) and the harmonic mean (H). VS, SS and SRS represents classifiers in visual, semantic, and searched representation space, respectively.}
	\label{table:AblationGZSL}
	\begin{tabular}{p{1cm}|p{2.9cm}|p{2.2cm}|p{0.3cm}p{0.3cm}p{0.3cm}|p{0.3cm}p{0.3cm}p{0.3cm}|p{0.3cm}p{0.3cm}p{0.3cm}|p{0.3cm}p{0.3cm}p{0.3cm}}
		\hline
		\multirow{3}{1cm}{Variant}&\multirow{3}{2.9cm}{ Loss}&\multirow{3}{2.2cm}{Classifier}&\multicolumn{12}{|c}{Generalized Zero-Shot Learning}\\
		\cline{4-15}
		& & & \multicolumn{3}{|c}{AWA1}&\multicolumn{3}{|c}{CUB}&\multicolumn{3}{|c}{APY}&\multicolumn{3}{|c}{SUN}\\
		\cline{4-15}
		& & &\multicolumn{1}{c}{U}&\multicolumn{1}{c}{S}&\multicolumn{1}{c|}{H}&\multicolumn{1}{c}{U}&\multicolumn{1}{c}{S}&\multicolumn{1}{c|}{H}&\multicolumn{1}{c}{U}&\multicolumn{1}{c}{S}&\multicolumn{1}{c|}{H}&\multicolumn{1}{c}{U}&\multicolumn{1}{c}{S}&\multicolumn{1}{c}{H}\\
		\hline
		\multirow{1}{1cm}{A}&\tabincell{c}{$\mathcal{L}_{WGAN}+\mathcal{L}_{F1}$\\($\mathbf{x}_{fake} = G(\mathbf{a},\mathbf{z})$)} & VS&
		55.1&64.7&59.5& 30.5&51.0&38.2& 21.6&55.2&31.1& 34.1&37.3&35.6\\
		\hline
		\multirow{1}{1cm}{B}&\tabincell{c}{$\mathcal{L}_{WGAN}+ \mathcal{L}_{F1}+\mathcal{L}_{F2}$\\($\mathbf{x}_{fake} = G(\mathbf{a},R(\mathbf{a}),\mathbf{z})$)}& VS&
		58.5&62.0&60.2& 38.5&50.7&43.8& 22.7&54.0&32.0& 36.3&35.1&35.7\\
		\hline
		\multirow{5}{1cm}{C}& \multirow{5}{3cm}{$\mathcal{L}_{DCRGAN} = \mathcal{L}_{WGAN-SR} + \mathcal{L}_{F1}+\mathcal{L}_{F2}$}&(C-1) VS
		&54.1&64.7&58.9& 44.0&54.6&48.7& 29.8&60.6&39.9& 40.5&36.0&38.1\\
		\cline{3-15}
		& &(C-2) SS
		&56.3&\textbf{75.7}&64.6& 54.4&34.4&42.1& 31.1&44.9&36.8& 48.0&8.6&15.0\\
		\cline{3-15}
		& &(C-3) SRS
		&49.9&62.5&57.1& 33.6&27.3&30.1& 32.3&71.1&44.4& 24.0&9.8&14.0\\
		\cline{3-15}
		& &(C-4) VS+SRS
		&56.1&65.9&60.6 &47.1&54.8&50.6& \textbf{37.9}&66.3&48.2& 42.1&36.0&38.8\\
		\cline{3-15}
		& &(C-5) VS+SS+SRS
		&\multicolumn{1}{c}{\textbf{62.7}} & \multicolumn{1}{c}{73.3} & \multicolumn{1}{c|}{\textbf{67.6}} & \multicolumn{1}{|c}{\textbf{55.8}} & \multicolumn{1}{c}{\textbf{66.8}} & \multicolumn{1}{c|}{\textbf{60.8}} & \multicolumn{1}{|c}{37.2} & \multicolumn{1}{c}{\textbf{71.7}} & \multicolumn{1}{c|}{\textbf{49.0}} & \multicolumn{1}{|c}{\textbf{47.1}} & \multicolumn{1}{c}{\textbf{38.5}} & \multicolumn{1}{c}{\textbf{42.4}} \\
		\hline
	\end{tabular}
\end{table*}

We compare our model with the recent state-of-the-arts published in the last few years.
Embedding methods include DEVISE \cite{frome2013devise} (NeurIPS13), DAP \cite{lampert2014attribute} (TPAMI14), SSE \cite{zhang2015zero} (ICCV15), SJE \cite{akata2015evaluation} (CVPR15), ESZSL \cite{bernardino2015an} (ICML15), ALE \cite{akata2016label} (TPAMI16), LATEM  \cite{xian2016latent} (CVPR16), SYNC \cite{changpinyo2016synthesized} (CVPR16), SAE \cite{kodirov2017semantic} (CVPR17), CRNet \cite{zhang2019co} (ICML19) and DVBE \cite{min2020domain} (CVPR20);
generative methods include GAZSL (CVPR18) \cite{zhu2018a}, PSR (CVPR18) \cite{annadani2018preserving}, f-CLSWGAN \cite{xian2018feature} (CVPR18), CDL \cite{jiang2018learning} (ECCV18), SRGAN \cite{ye2019srgan} (ICME19), GDAN \cite{huang2019generative} (CVPR19), DASCN \cite{ni2019dual} (NeurIPS19), AFC-GAN \cite{li2019alleviating} (ACM MM19), OCD-VAE \cite{keshari2020generalized} (CVPR20), EUC-VAE \cite{chen2021entropy} (arxiv21) and LsrGAN \cite{vyas2020leveraging} (ECCV20).
The results of ZSL and GZSL are reported in Table~\ref{table:OverallComparison}.
We first take an overall comparison with the state-of-the-arts in Section~\ref{sec:gzsl} for ZSL and GZSL.
Then, we compare our approach with others in Section~\ref{sec:ganbased} and Section~\ref{sec:tlbased} from two aspects, GAN-based and triplet-loss-based viewpoints, respectively.

\subsubsection{(Generalized) Zero-shot Learning}
\label{sec:gzsl}
For ZSL, from the results reported in Table~\ref{table:OverallComparison}, we can see that we obtain 2.5\%, 0.1\% improvements on APY and SUN, respectively, against the previous state-of-the-art.

For GZSL, we follow previous work~\cite{xian2017zero} and report the harmonic mean that can avoid the effects of extreme values.
For instance, we can see from the results that SAE gets 1.8\% for unseen and 77.1\% for seen on CUB.
Although the accuracy of seen classes is the best, the harmonic mean is only 3.5\% due to the extremely low result on unseen categories.
In a nutshell, the harmonic mean is high only if the accuracies on both seen and unseen categories are high.
From the results, we can observe that our method achieves overall the best harmonic mean on all of the evaluations besides SUN.
It indicates that our DCR-GAN is a stable method which can work well for both seen and unseen instances.
In details, DCR-GAN outperforms the state-of-the-art LsrGAN with $4.6\%$, $7.8\%$ and $1.5\%$ improvements on AWA1, CUB and SUN, respectively.
Notably, our method achieves the best on AWA1, CUB of the unseen categories, which significantly outperforms AFC-GAN by 4.5\% and 2.3\%.
It is worth mentioning that we do not use any explicit constraint to avoid ``train bias" problem, but our proposed model can still surpass AFC-GAN that uses the boundary loss to compel synthesized unseen features to be far away from seen features.

\subsubsection{Comparison with GAN-based Approaches}
\label{sec:ganbased}
The GAN-based methods, f-CLSWGAN, cycle-CLSWGAN, AFC-GAN, GAZSL, SRGAN and LsrGAN, share the same basic loss $\mathcal{L}_{WGAN}$.
Benefiting from the prior semantic features to generate missing data, these GAN-based methods can obtain better performance than those earlier embedding approaches though one recent embedding method, i.e. DVBE also demonstrates excellent performance. In addition, GDAN is a new method that unifies generative, embedding, and metric learning as a basic architecture .
Benefiting from the change in the basic architecture, GDAN demonstrates  an excellent score $S$ for seen  classification in SUN, which indicates a better basic loss may be necessary.

Our approach DCR-GAN adopts the proposed  loss $\mathcal{L}_{WGAN-SR}$.
Comparing our DCR-GAN with other GAN-based methods, we observe that our method leads to competitive performance with others both in ZSL and GZSL.
More concretely, just SRGAN and our DCR-GAN can  recognize more than $70\%$ unseen samples of AWA1 in ZSL. Although DCR-GAN cannot beat GDAN on SUN in the  GZSL setting,  it attains $63.7\%$ and shows much better performance than GDAN and all the other methods on SUN for ZSL. These outstanding performance of our DCR-GAN demonstrates the effectiveness of our proposed basic loss $\mathcal{L}_{WGAN-SR}$.

It is noted that GDAN appears to perform much better than all the other methods including DCR-GAN on SUN in terms of $H$. 
Exploiting an adversarial loss $\mathcal{L}_{GDAN}$ to train the model,  it is observed that GDAN may overfit seen samples almost on all the  datasets. This results in its excellent performance $S$ for seen classes (particularly $S = 89.9\%$ in SUN), while the accuracy of $U$ appears much worse: lower than almost all the other GAN-based methods. Such drawback may limit its application in ZSL/GZSL where recognition of unseen samples may be more crucial. In contrast, our proposed adversarial  loss~$\mathcal{L}_{WGAN-SR}$ appears to achieve an outstanding balance for both seen and unseen classes.
As a matter of fact, our method outperforms GDAN in CUB and APY in both $U$ and $H$; it is also better than GDAN in terms of $U$ in SUN.

\subsubsection{Comparison with Triplet-loss-based Approaches}
\label{sec:tlbased}
Besides our DCR-GAN, the methods OCD-VAE and EUC-VAE also introduce TL for searching discriminative latent features. Our DCR-GAN exploits the proposed MMTL, while the other methods use the traditional single-modal TL. As observed, for ZSL, DCR-GAN outperforms the others on APY and SUN;
for GZSL,  the proposed method demonstrates the best performance on AWA1, CUB and APY.
It is noted that,  although DCR-GAN performs not as well as them on SUN for GZSL, our model generates better performance  for unseen recognition.
Namely, our DCR-GAN recognizes $47.1\%$ unseen samples in SUN, while OCD-VAE and EUC-VAE only recognize $44.8\%$ and $35.0\%$, respectively.

\begin{figure*}[htbp]
	\centering
	\includegraphics[width=\linewidth]{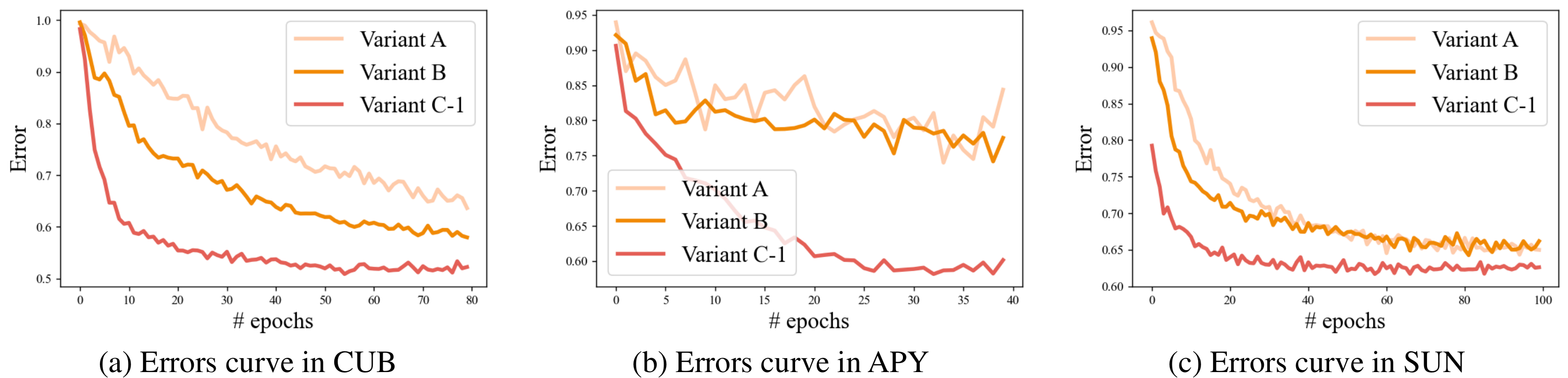}
	\caption{Comparison between the visual softmax classifiers from DCRGAN and two Variants.}
	\label{fig:ErrorCurve}
\end{figure*}

\subsection{Ablation Study}
\label{section:ablation}
To further verify the effectiveness of our approach, we take an ablation study on the searched class representations.
Table~\ref{table:AblationGZSL} reports three variants in the setting of GZSL, respectively.
We also provide convergence curve of three visual classifiers in Fig.~\ref{fig:ErrorCurve}.
The variant $A$ indicates the traditional WGAN-based ZSL method trained by:
\begin{equation}
	\mathcal{L}_{A} = \mathcal{L}_{WGAN} + \lambda_{1} \mathcal{L}_{F1},
	\label{eq:A}
\end{equation}
where $\mathbf{x}_{fake} = G(\mathbf{a},\mathbf{z})$.
To fairly verify the effectiveness of the searched representations, we also train the variant $B$.
We only change $\mathbf{x}_{fake}$ as $G(\mathbf{a},R(\mathbf{a}),\mathbf{z})$, and add in its reconstruction loss.
In a word, the variant $B$ is trained by:
\begin{equation}
\mathcal{L}_{B} = \mathcal{L}_{WGAN} + \lambda_{1} \mathcal{L}_{F1} + \lambda_{2} \mathcal{L}_{F2},
\label{eq:B}
\end{equation}
where $\mathbf{x}_{fake} = G(\mathbf{a},R(\mathbf{a}),\mathbf{z})$.
Another variant $C$ presents the complete  DCR-GAN model trained by $\mathcal{L}_{DCRGAN}$.
VS, SS and SRS represents classifiers in the visual, semantic and searched representation space, respectively.
Comparing the results, we highlight our discussion as follows:
\begin{enumerate}
\item[(1)] \textbf{Effectiveness of our searched representation}: Comparing Table~\ref{table:AblationGZSL} A and B, we can find that the harmonic mean accuracies $H$ are significantly improved in both CUB and APY;  unseen accuracies $U$ are  greatly improved on AWA1 and SUN. These results show that our searched representations can help $G$ to fit more realistic distributions.

Then, comparing the variants C-1 and C-4, we can find that accuracies $H$ in all the datasets are improved by integrating the classification results in the searched representation space.
Specifically, $H$ increases most notably in APY, from $39.9\%$ to $48.2\%$.
Even in the most indistinctive dataset SUN, our classifier SRS  can still bring $1.6\%$ improvement for unseen recognition.

\item[(2)] \textbf{Effectiveness of our $\mathbf{\mathcal{L}_{WGAN-SR}}$}:
The variant B and C-1 are different only  in terms of the generation loss.
B is trained by $\mathcal{L}_{WGAN}$, while C-1 is trained by our proposed $\mathcal{L}_{WGAN-SR}$. Inspecting Table~\ref{table:AblationGZSL} B and C-1, we can find that  though $H$ is slightly decreased in AWA1, $H$ is increased  notably in the remained three datasets.
For example, in the mid-scale dataset SUN, $H$ is lifted up  by $2.4\%$ (from $35.7\%$ to $38.1\%$).
These results validate  the effectiveness of our proposed $\mathcal{L}_{WGAN-SR}$.
\end{enumerate}

\begin{figure}[htbp]
	\centering
	\includegraphics[width=0.97\linewidth]{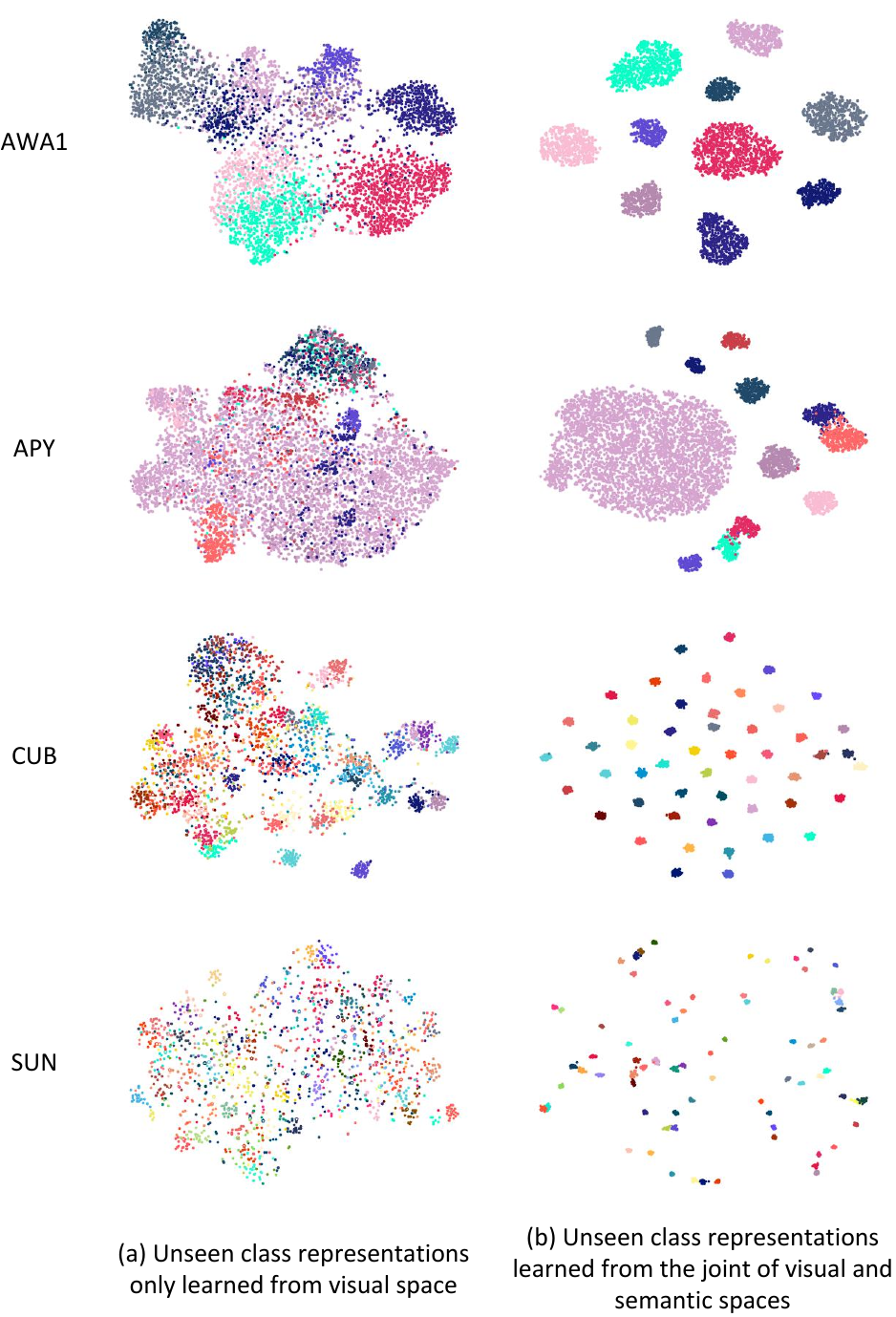}
	\caption{Visualization of unseen class representations.
		 (a) is searched by the traditional TL (Eq.~\ref{eq:TL});  (b) is searched by the MMTL (Eq.~\ref{eq:MMTL}). Different colors indicate different classes.}
	\label{fig:vis2}
\end{figure}

\begin{figure*}[htbp]
	\centering
	\includegraphics[width=0.95\linewidth]{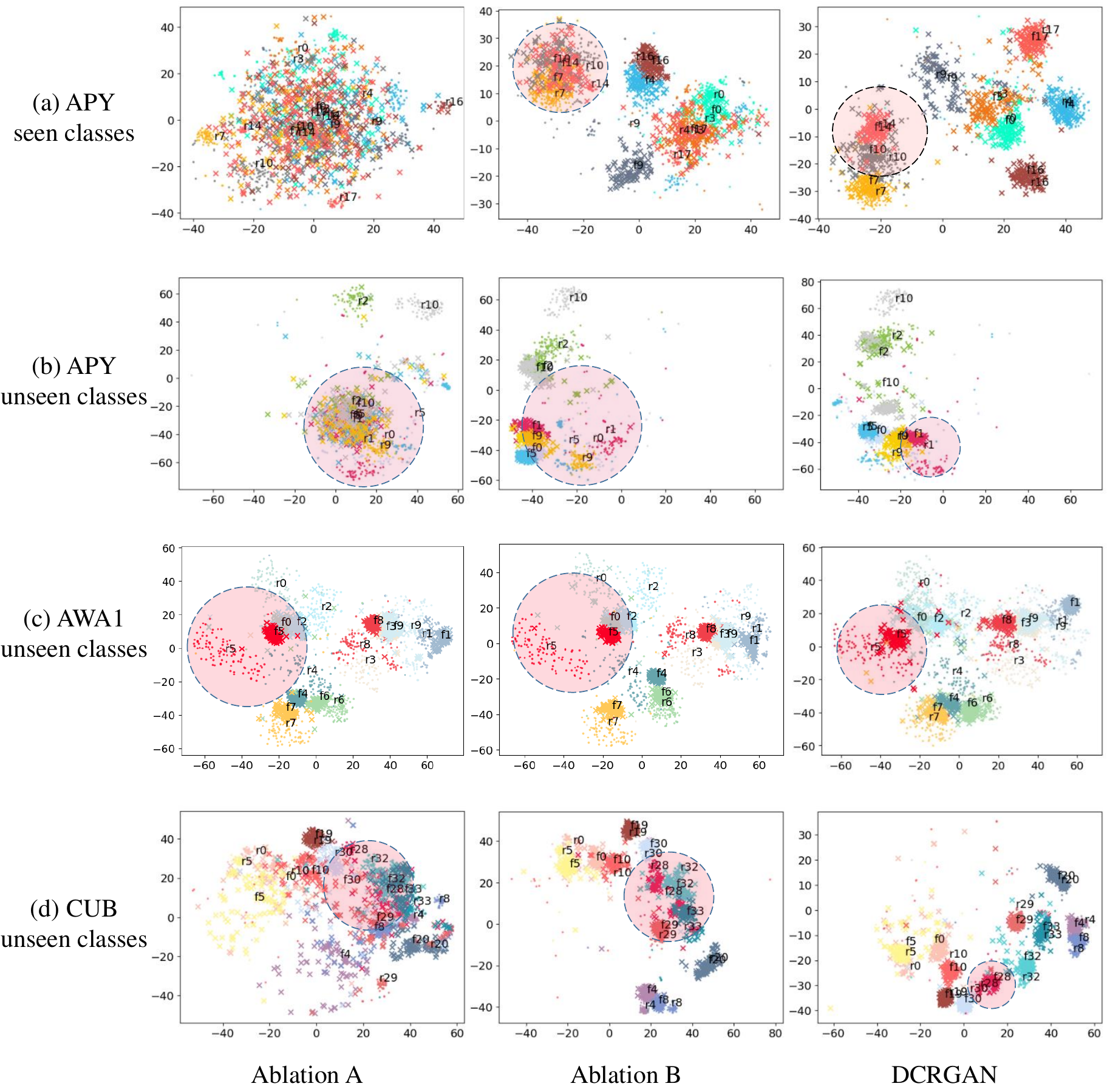}
	\caption{Visualization of synthesized visual features.
	Points denote real features and crosses denote synthesized features.
	Different colors and numbers indicate different classes.}
	\label{fig:fakefeature}
\end{figure*}


\subsection{Visualization of Feature Space}
\label{section:vis}
In order to provide a qualitative evaluation on our proposed DCR-GAN, we first visualize two kinds of unseen class representations as shown in Fig.~\ref{fig:vis2}, where (a) is one kind of augmented semantic learned by a triplet loss only from the visual space, i.e. the semantic augmentation method of LDF~\cite{li2018discriminative}.
Clearly, this augmented semantic is muddled.
Various categories are mixed in the representation space.
Fig.~\ref{fig:vis2} (b) shows the visualization of our searched class representations, which are learned both from visual and semantic spaces.
As observed, by utilizing the original semantic information, our model decouples searched class representations where boundaries between each category are clear.
Note that our model does \textit{not} use any unseen features in training.
This confirms our idea ---the original semantic information is also necessary in the augmented semantic searching process.

In addition, we visualize synthetic image features along with real image features.
These results are illustrated in Fig.~\ref{fig:fakefeature}.
Since the numbers of categories of CUB and SUN are too large to visualize, we only show the results of seen classes of APY and unseen classes of APY, AWA1 and CUB.
For each class we synthesize 100 features, and then we use t-SNE \cite{matten2018visualizaing} to reduce the dimension to two for visualization.
The synthesized features of $i$-th class are marked by f$i$, and real features are marked by r$i$.
It is evident that our searched class representations successfully help DCRGAN to synthesize more realistic visual features than those of the baseline.
Some synthesized features by our DCRGAN are almost the same as real features, e.g. $14$-th seen class in APY, $1$-st unseen class in APY, $5$-th unseen class in AWA1 and $28$-th of unseen classes in CUB.
This visualization again indicates the identified \emph{entangled unseen visual feature problem} problem.
More importantly, our searched class representation can help generative models to fit more realistic distribution.

\section{Conclusion}
In this paper, we argue that the \emph{entangled unseen visual feature problem} exists in the current Zero-shot Learning (ZSL) and Generalized ZSL.
We propose our generative framework DCR-GAN to address the problem.
DCR-GAN contains two novel loss: a multi-modal triplet loss (MMTL), and an adversarial loss $\mathcal{L}_{WGAN-SR}$.
Compared with the traditional TL, MMTL is capable of searching more decoupled unseen class representations.
Our designed $\mathcal{L}_{WGAN-SR}$ can reduce high risks of hard sample generation.
Benefiting from our MMTL and $\mathcal{L}_{WGAN-SR}$, our model learns more realistic distribution and generates more disentangled features.
With the searched class representations, our DCR-GAN synthesizes visual features from semantic features and searched class representation.
Given synthesized visual features, we train a softmax classifier for the visual space.
Additionally, we ensemble the semantic and searched class representation softmax with the visual one.
Experimental results show that the proposed approach achieves state-of-the-art performance on ZSL task and boosts the performance by a great margin for Generalized ZSL. 


%

\section*{Acknowledgment}

This work was supported by the National Natural Science Foundation of China (Nos. 61876121, 62002254, 61801323, 61876155), the Jiangsu Provincial Key Research and Development Program (Nos. BE2017663, BE2020006-4B), and the Natural Science Foundation of the Jiangsu Higher Education Institutions of China (No. 19KJB520054).

\ifCLASSOPTIONcaptionsoff
  \newpage
\fi



%
\bibliographystyle{IEEEtran}
\bibliography{refs}


%

\begin{IEEEbiography}[{\includegraphics[width=1in,height=1.25in,clip,keepaspectratio]{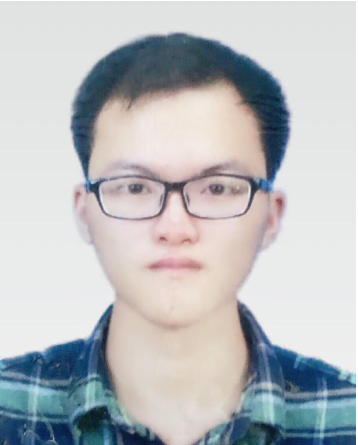}}]{Zihan Ye}
is an undergraduate student from Suzhou University of Science and Technology and a student member of the IEEE.
His research interests include zero-shot learning, generative adversarial network, computer vision, and deep learning.
He has served as a reviewer for IEEE SPL.
In 2020 Oct, he taked an oral presentation at IEEE International Conference on Image Processing (ICIP).
In 2019, he taked an oral presentation at IEEE International Conference on Multimedia \& Expo (ICME).
In 2018, he received the best poster paper award at The 8th International Conference on Brain Inspired Cognitive Systems (BICS).
(Email: zihhye@outlook.com)
\end{IEEEbiography}

\begin{IEEEbiography}[{\includegraphics[width=1in,height=1.25in,clip,keepaspectratio]{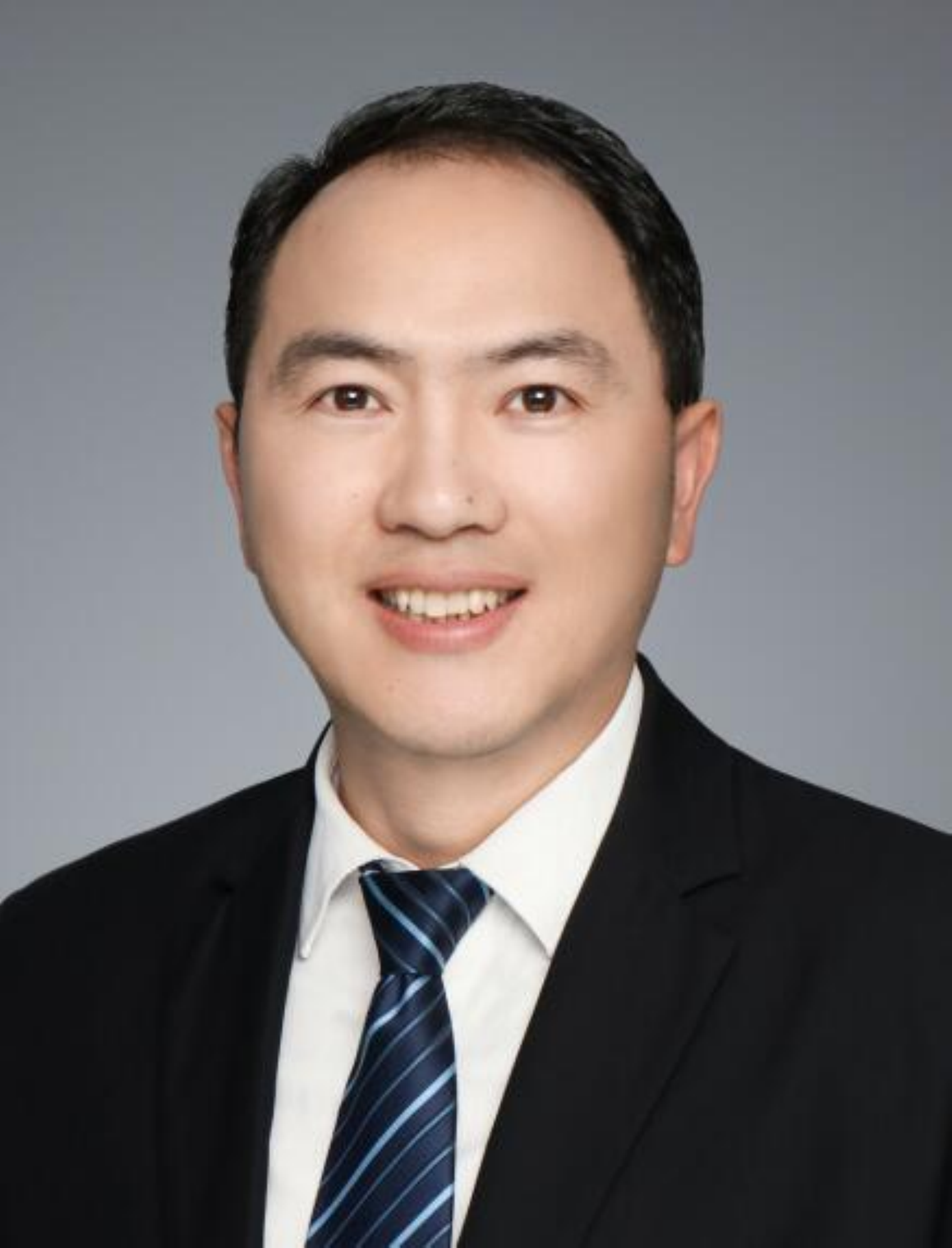}}]{Fuyuan Hu}
	(corresponding author) was a postdoctoral researcher at Vrije Universiteit Brussel, Belgium, a Ph.D. student at Northwestern Polytechnical University, and a visiting Ph.D. student at
	the City University of Hong Kong. He is a professor at Suzhou University of Science and Technology. His research interests include machine learning, graphical models, structured learning, and tracking. (Email: fuyuanhu@mail.usts.edu.cn)
\end{IEEEbiography}

\begin{IEEEbiography}[{\includegraphics[width=1in,height=1.25in,clip,keepaspectratio]{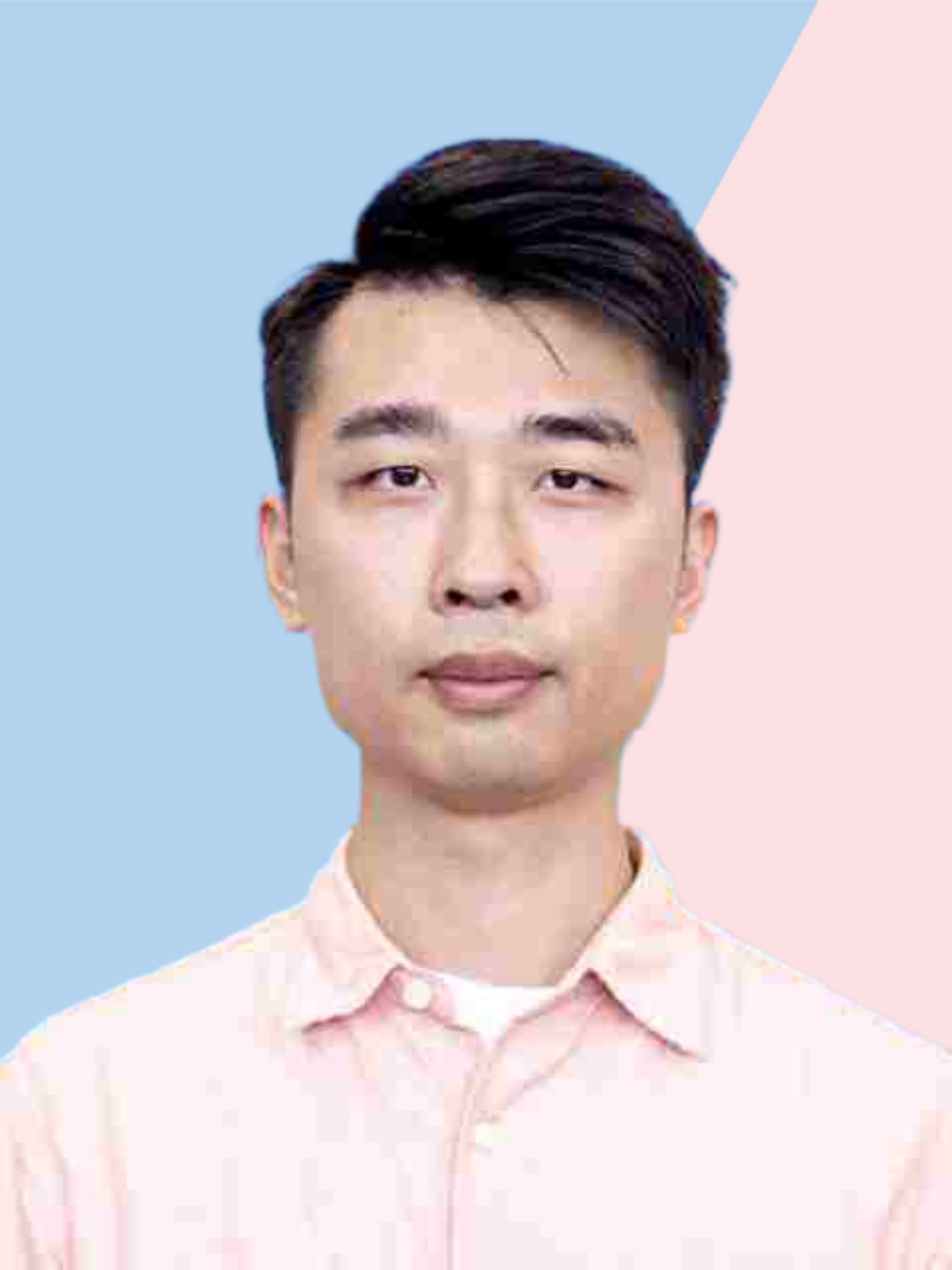}}]{Fan Lyu}
	is an undergraduate student from Suzhou University of Science and Technology and a student member of the IEEE.
	received the BS and MS degree in Electronic \& Information Engineering, Suzhou University of Science and Technology, China, in 2015 and 2018. He is working toward the PhD degree in the College of Intelligence and Computing, Tianjin University, China. His research interests include lifelong learning, vision-language understanding, low-shot learning, etc.
\end{IEEEbiography}

\begin{IEEEbiography}[{\includegraphics[width=1in,height=1.25in,clip,keepaspectratio]{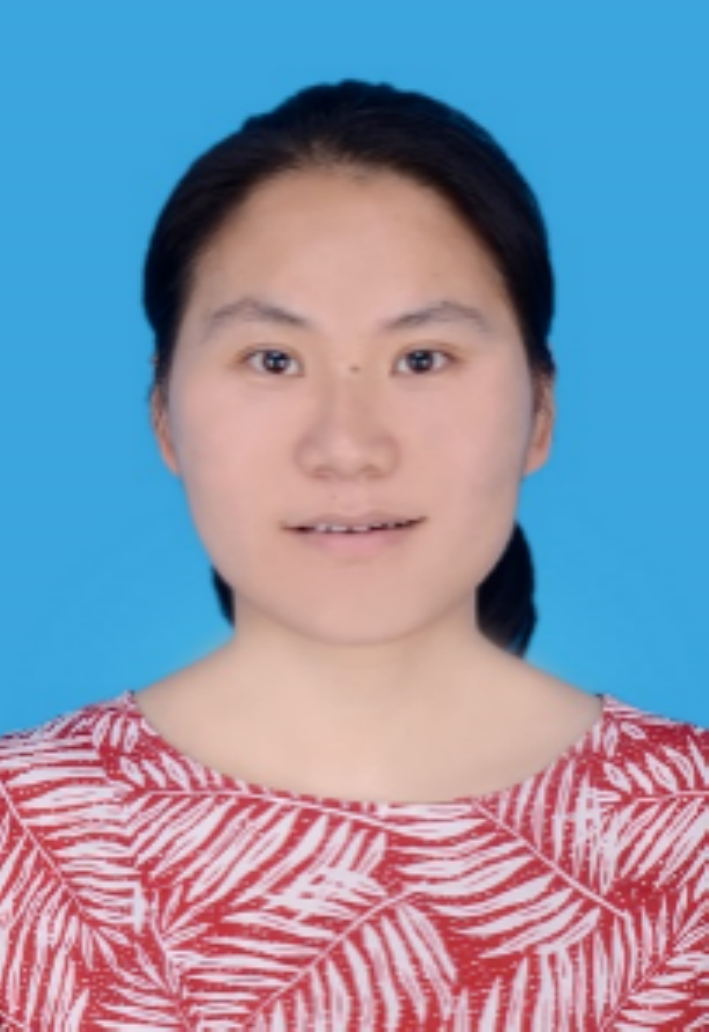}}]{Linyan Li}
    is currently an associate Professor at Suzhou Institute of Trade \& Commerce, China. She obtained her Master degree from Wuhan University in 2007.She has been working in machine learning, neural information processing, and pattern recognition.
\end{IEEEbiography}

\begin{IEEEbiography}[{\includegraphics[width=1in,height=1.25in,clip,keepaspectratio]{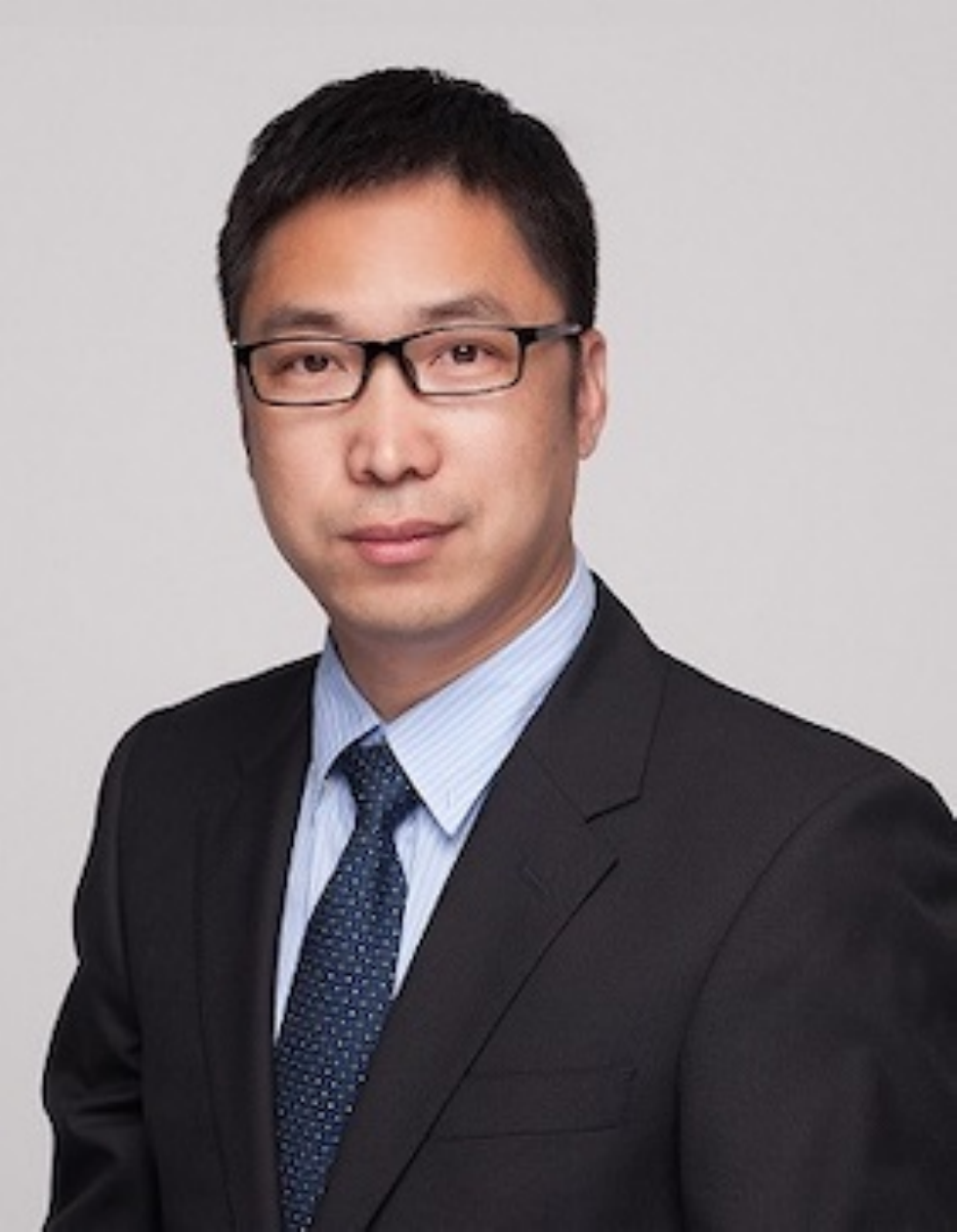}}]{Kaizhu Huang}
	 (corresponding author) is currently a Professor at Xi’an Jiaotong-Liverpool University, China. Prof. Huang obtained his PhD degree from Chinese University of Hong Kong (CUHK) in 2004. He worked in Fujitsu Research Centre, CUHK, University of Bristol, National Laboratory of Pattern Recognition, Chinese Academy of Sciences from 2004 to 2012. Prof. Huang has been working in machine learning, neural information processing, and pattern recognition. He was the recipient of 2011 Asia Pacific Neural Network Society Young Researcher Award. He received best paper or book award five times. Until September 2020, he has published 9 books and over 190 international research papers (70+ international journals) e.g., in journals (JMLR, Neural Computation, IEEE T-PAMI, IEEE T-NNLS, IEEE T-BME, IEEE T-Cybernetics) and conferences (NeurIPS, IJCAI, SIGIR, UAI, CIKM, ICDM, ICML, ECML, CVPR). He serves as associated editors/advisory board members in a number of journals and book series. He was invited as keynote speaker in more than 20 international conferences or workshops. (E-mail: Kaizhu.Huang@xjtlu.edu.cn).
\end{IEEEbiography}




\end{document}